\documentclass{article}

     \PassOptionsToPackage{numbers, sort&compress}{natbib}
\usepackage{graphicx, color}
\usepackage[utf8]{inputenc}
\usepackage[english]{babel}
\usepackage{amsfonts}
\usepackage{xcolor}
\usepackage{graphicx}
\usepackage{subcaption}
\usepackage{caption}      % optional: fine-tune caption layout
\usepackage{amsmath}      % for math in caption
\usepackage{wrapfig}
\usepackage{hyperref}
\hypersetup{
    colorlinks=true,
    citecolor=gray,
    breaklinks=true,
}% hyperlinks

\usepackage{booktabs}
\usepackage{float}
\usepackage{tocloft}

\usepackage{iclr2026_delta,times}
\iclrfinalcopy

\title{Dynamic Mixture-of-Experts for Visual Autoregressive Model\thanks{Code is available at \url{https://github.com/JortVincenti/DMoE-VAR}.}}

\usepackage[utf8]{inputenc} % allow utf-8 input
\usepackage[T1]{fontenc}    % use 8-bit T1 fonts
\usepackage{hyperref}       % hyperlinks
\usepackage{url}            % simple URL typesetting
\usepackage{booktabs}       % professional-quality tables
\usepackage{amsfonts}       % blackboard math symbols
\usepackage{nicefrac}       % compact symbols for 1/2, etc.
\usepackage{microtype}      % microtypography
\usepackage{xcolor}         % colors
\usepackage{enumitem}
\author{Jort Vincenti$^{1}$, Metod Jazbec$^{1}$, Guoxuan Xia$^{2}$ \\
$^{1}$University of Amsterdam, UvA-Bosch Delta Lab
$^{2}$Imperial College London}

\begin{document}

\maketitle
\vspace{-2.5ex} 

\begin{abstract}
    Visual Autoregressive Models (VAR) offer efficient and high-quality image generation but suffer from computational redundancy due to repeated Transformer calls at increasing resolutions. We introduce a dynamic Mixture-of-Experts router integrated into VAR. The new architecture allows to trade compute for quality through scale-aware thresholding. This thresholding strategy balances expert selection based on token complexity and resolution, without requiring additional training. As a result, we achieve $\sim\!20\%$ fewer FLOPs, $\sim\!11\%$ faster inference and match the image quality achieved by the dense baseline. 
\end{abstract}
\vspace{-2.5ex}
\section{Introduction} 

Autoregressive Modelling has not matched diffusion models in image generation quality despite demonstrating strong performance in text \cite{ma2025}. The Visual Autoregressive Model (VAR) \cite{tian2024visualautoregressivemodelingscalable} bridges this gap by shifting from next-token to next-scale prediction, iteratively refining images by invoking an autoregressive transformer to predict each subsequent resolution. VAR is the first model to reach diffusion models in image quality while also improving runtime.

A key goal in autoregressive modeling is improving computational efficiency without sacrificing performance. Dynamic methods address this by allocating computation based on token complexity \cite{graves2016adaptive}. Currently, VAR still makes multiple static transformer calls per scale, ignoring varying complexity across tokens and scales. Exploiting these variations could enhance efficiency by focusing computation on complex regions and progressively reducing resources at finer scales \cite{rao2021dynamicvit}. Dynamic allocation at the scale level remains unexplored, presenting an opportunity to improve VAR’s efficiency. As a result, we adapt a dynamic Mixture-of-Experts (MoE) framework that routes experts based on token complexity, specifically for VAR’s coarse-to-fine structure \cite{szatkowski2024exploitingactivationsparsitydense}. We make the following contributions:

\begin{enumerate}[topsep=1ex,leftmargin=1.5em]
    \item A variant of the VAR model where each feed-forward block is replaced by a Mixture-of-Experts layer with an adaptive router, enabling the model to allocate resources across tokens and scales based on their complexity.
    \vspace{-0.5ex}
    \item An inference-tunable and scale-dependent threshold that activates fewer experts at higher resolutions, matching the observation that fine-scale tokens need less computation.
    \vspace{-0.5ex}
    \item An empirical validation on ImageNet \cite{deng2009imagenet} showing 19\% fewer FLOPs and 11\% faster wall-clock time while staying within 1\% Fréchet Inception Distance score \cite{heusel} of the VAR depth 16 baseline. 
\end{enumerate}

\vspace{-1ex}
\section{Preliminaries}\label{chap:preliminaries}

\subsection{Visual Autoregressive Modelling (VAR)}\label{sec:prelim:var}

Rather than predicting tokens sequentially, VAR \cite{tian2024var} redefines image generation as a coarse-to-fine process across multiple resolutions. An image is represented as a hierarchy of discrete token maps, ranging from a low to high resolution. The model then learns to autoregressively predict each finer scale conditioned on the coarser ones, progressively refining structure and texture until the full image is reconstructed. While this hierarchical design improves efficiency compared to raster-scan autoregression, related works for accelerating inference in VAR is discussed in Appendix \ref{section:efficientInference}.

\textbf{Stage 1: Multi-scale VQ-VAE encoder.} VAR employs a multi-scale VQ-VAE to discretize images into a sequence of token maps at progressively coarser resolutions. The process starts from the full-resolution image, which is encoded into a dense feature map. This feature map is then quantized by assigning each vector to its nearest entry in a shared codebook, producing one map of tokens. This procedure is repeated multiple times, until the entire image is summarized by a single token. The result is a hierarchy of token maps which provide the discrete multi-scale representation used in Stage 2.

\textbf{Stage 2: Next-scale prediction.} After obtaining the token hierarchy $\mathbf{R}=(\mathbf{r}_1,\dots,\mathbf{r}_K)$, a Transformer is trained autoregressively to predict each scale conditioned on previous scales. During training, the Transformer takes the sequence $(\mathtt{[s]}, \mathbf{r}_1, \dots, \mathbf{r}_{K-1})$ as input and employs a block-wise causal attention mask, ensuring that tokens at scale $k$ can only attend to earlier scales. At inference, tokens are generated recursively from coarsest ($\mathbf{r}_1$) to finest ($\mathbf{r}_K$), after which a VQ-VAE decoder reconstructs the final image. While generation at each scale occurs in parallel, the computational cost increases significantly with higher resolutions due to the tokens growing quadratically from just one token at $k{=}1$ to 256 at $k{=}K$.

\subsection{Dense-to-Dynamic-$k$ Mixture-of-Experts (D2DMoE)}\label{sec:prelim:d2dmoe}

\textbf{From dense FFN to experts.} In each Transformer block the two-layer FFN computes the hidden vector $\mathbf{h}=\sigma(\mathbf{W}_1\mathbf{x}+\mathbf{b}_1)$ and outputs $F(\mathbf{x})=\mathbf{W}_2\mathbf{h}+\mathbf{b}_2$. To make the activations $\mathbf{h}$ even sparser than with $\sigma$ set to ReLU, D2DMoE adds a Hoyer penalty \cite{hoyer2004} to the cross entropy loss:
\vspace{-0.5ex}
\begin{equation}\label{eq:losses}
\begin{aligned}
L(x)   &= L_{\mathrm{CE}}(\hat{y},y)+\alpha\,L_s(x), \qquad
L_s(x) &= \frac{1}{L}\sum_{\ell=1}^{L}
          \frac{(\sum_i |h_i^{(\ell)}|)^2}{\sum_i (h_i^{(\ell)})^2}
\end{aligned}
\end{equation}
\vspace{-0.5ex}
\noindent where $\mathbf{h}^{(\ell)}$ is the post-activation vector of FFN block $\ell$, $L$ is the number of blocks, and $\alpha$ controls sparsity strength. After sparsity-aware training, columns of $\mathbf{W}_1$ respond to subsets of tokens. By grouping these columns via $k$-means clustering into balanced clusters \cite{Zhang2022MoEfication}, D2DMoE rearranges the original FFN parameters into multiple distinct experts creating a Mixture-of-Experts layer.

\textbf{Regression-based routing.} D2DMoE introduces a routing mechanism framed as a regression task rather than the traditional top-$k$ selection. A lightweight router $R$ predicts the $\ell_2$-norm of each expert’s output for an input token $\mathbf{x}$, minimising the difference between predicted and true norms via: 
\vspace{-0.5ex}
\begin{equation}
\label{routernorm}
  L_{r}(x) = \frac{1}{n} \sum_{i=1}^{n} \big(R(x)_i - \|E_i(x)\|\big)^2
\end{equation} 
At inference, experts are dynamically selected using a relative thresholding mechanism. An expert would be selected only if $R(x)_{i} \geq \tau \cdot \max(R(x))$, where the adjustable scalar $\tau \in [0,1]$ provides flexible control over computational cost and performance without needing to retrain.

\section{Dynamic Mixture-of-Experts for Visual Autoregressive Model}\label{sec:divar}

\textbf{Expert Construction.} As detailed in section \ref{sec:prelim:d2dmoe}, at every Transformer layer, we replace the standard VAR's FFN with a MoE block. Experts are built offline by clustering the sparse FFN weights from the ReLUfied model \cite{relustrikesback} finetuned with the Hoyer-norm regularisation \eqref{eq:losses}. We then apply the same equation as in \eqref{routernorm} to train a router that regresses the $\ell_2$-norm of each expert.

\textbf{$\tau$-Based Expert Selection.} At inference, we route tokens through a stack of $S$ progressively finer scales, with each scale $s = 1,\dots,S$ indexed by increasingly thresholds $\tau_1 < \tau_2 < \dots < \tau_S$. At each scale $s$, the router outputs a vector of norm predictions $R^{s}(x) \in \mathbb{R}^E$ across the $E$ experts. Expert selection is then determined by applying the following rule to each expert $i$:
\vspace{-0.5ex}
\begin{equation}
    G^{s}(x)_i =
    \begin{cases}
    1 & \text{if } R^{s}(x)_i \ge \tau_s \max_j R^{s}(x)_j \\
    0 & \text{otherwise}
    \end{cases}
\label{dynamicthresh}
\end{equation}

\begin{figure}[htbp]
  \centering
\includegraphics[width=1.0\linewidth]{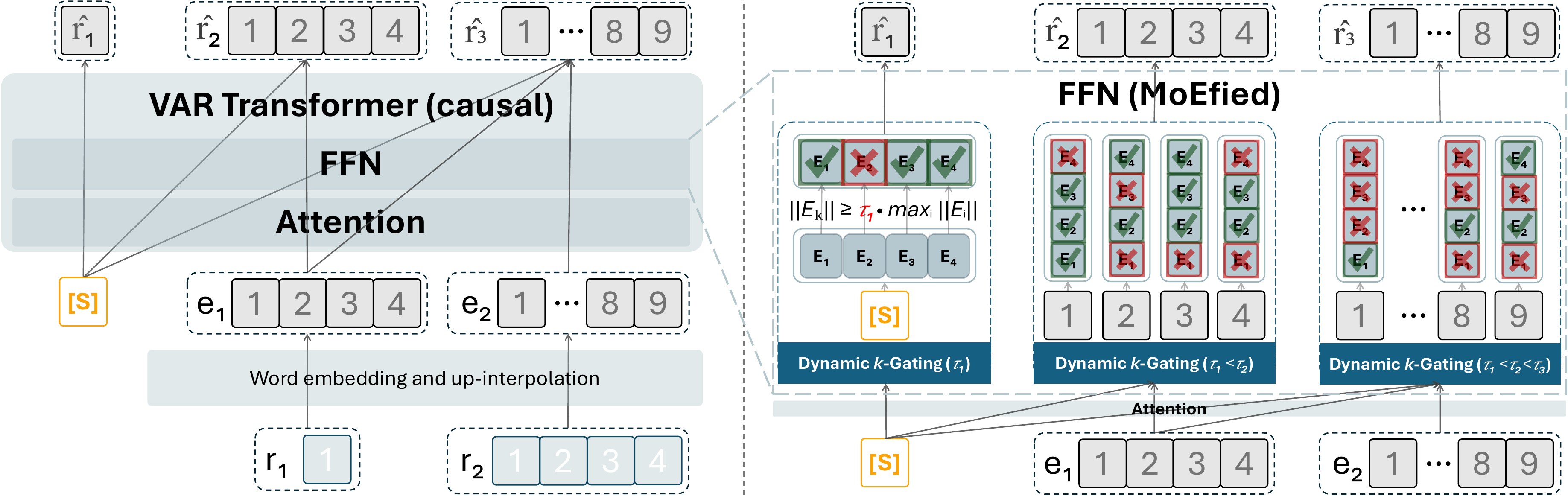}
\caption{\textbf{Generation pipeline.} \emph{Left:} The coarse-to-fine decoder of VAR performing next-scale prediction: it takes $(\mathtt{[s]},r_1,r_2,\dots,r_{K-1})$ as input to predict $(\hat{r}_1,\hat{r}_2,\dots,\hat{r}_K)$. \emph{Right:} The FFN block is replaced with a dynamic-$k$ gating MoE module. It executes expert $E_j$ only when $\|E_j\|\!\ge\!{\color{red}\tau_s}\max_i\|E_i\|$, filtering the experts by $\ell_2$-norm. Because the thresholds grow with resolution $\tau_{1}<\dots<\tau_{S}$, many experts are used at coarse scales while fewer are consulted at fine scales.}
\label{fig:DIVARMOE}
\vspace{-3ex}
\end{figure}

\noindent where $G^{s}(x)_i$ indicates whether expert $i$ receives the token $x$ at scale $s$. This $\tau$-based selection ensures that compute allocation is both token- and scale-aware: lower thresholds at coarse scales admit more experts, while higher thresholds at finer scales enforce greater sparsity with fewer selected experts (as illustrated in Fig.~\ref{fig:DIVARMOE}).

Overall, our method unifies the coarse-to-fine paradigm of VAR with the adaptive computational load of the MoE framework. By merging routing granularity (token-level) with resolution granularity (scale-level), we achieve:
\vspace{-0.5ex}
\begin{enumerate}[topsep=0ex,leftmargin=1.5em]
    \item \textbf{Scale-aware budget scheduling} via $\tau_{k}$, whereby each successive scale requires less compute.
    \item \textbf{Token-aware compute allocation}, which is crucial for handling the highly skewed sparsity patterns within larger scales.
\vspace{-1ex}
\end{enumerate}

\section{Results}
\vspace{-0.5ex}
\textbf{Implementation Details.} All experiments use the pre-trained VAR-d16 baseline, fine-tuned for two epochs with the sparse cross-entropy loss in Eq.~\eqref{eq:losses}. The full set of hyperparameters is provided in Appendix~\ref{Appendix:Hyperparameter}. All experiments are conducted on NVIDIA A100 GPUs, measuring floating-point operations (FLOPs) using \texttt{fvcore}\footnote{\url{https://github.com/facebookresearch/fvcore}} and Fréchet Inception Distance (FID) using \texttt{torch-fidelity}\footnote{\url{https://github.com/toshas/torch-fidelity}}, with evaluation on ImageNet using 10{,}000 samples~\cite{deng2009imagenet}.

\textbf{Design Exploration.} Our design exploration in Appendix~\ref{Section:DesignExploration} provides several key insights. First, we find that Hoyer regularization mainly induces sparsity in early scales, contributing little to computational savings, while ReLUfication yields more substantial sparsity overall (Appendix~\ref{Section:Sparsity}). Second, practical speed-ups require configuring fewer, larger experts and limiting MoEfication to later scales (Appendix~\ref{Section:MoEfication}). Finally, routers with higher sparsity regularisation ($\alpha\!=\!0.1$) achieve better FID, likely due to simpler routing from zero-valued expert norms (Appendix~\ref{Section:RouterAndTau}). These findings motivate our final design which fine-tunes the ReLUfied model with Hoyer sparsity ($\alpha\!=\!0.1$), then applies MoE layers using 32 experts of size 128, using the model only on the last three scales.

\begin{figure}[!ht]
  \centering
  \setlength{\tabcolsep}{0pt}\renewcommand{\arraystretch}{0}
  \begin{subfigure}[t]{0.39\linewidth}
    \vspace{0pt}
    \centering
    {\small\textbf{VAR}}\\[2pt]
    \begin{tabular}{@{}cccc@{}}
      \includegraphics[width=0.24\linewidth]{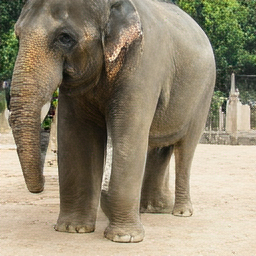} &
      \includegraphics[width=0.24\linewidth]{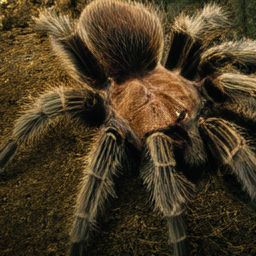} &
      \includegraphics[width=0.24\linewidth]{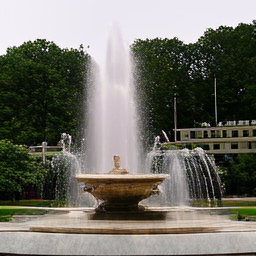} &
      \includegraphics[width=0.24\linewidth]{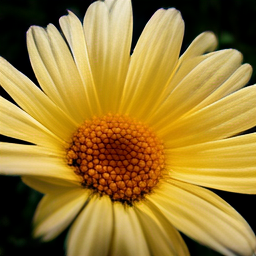} \\
    \end{tabular}
    
    \vspace{0.6em}
    
    {\small\textbf{DMoE--VAR (ours)}}\\[2pt]
    \begin{tabular}{@{}cccc@{}}
      \includegraphics[width=0.24\linewidth]{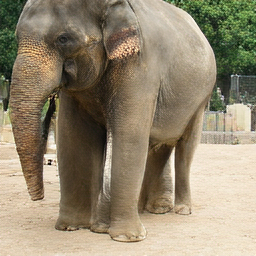} &
      \includegraphics[width=0.24\linewidth]{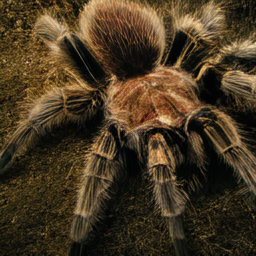} &
      \includegraphics[width=0.24\linewidth]{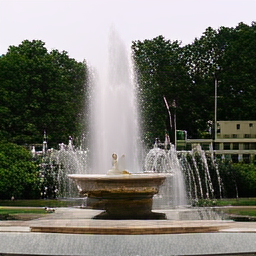} &
      \includegraphics[width=0.24\linewidth]{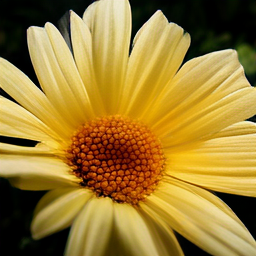} \\
    \end{tabular}
    \caption{Image comparison}
    \label{fig:imgcomparaisons}
  \end{subfigure}
  % LEFT (trim extra margins from the PDF)
  \begin{subfigure}[t]{0.6\linewidth}
    \vspace{0pt}%
    \centering
    \includegraphics[width=\linewidth]{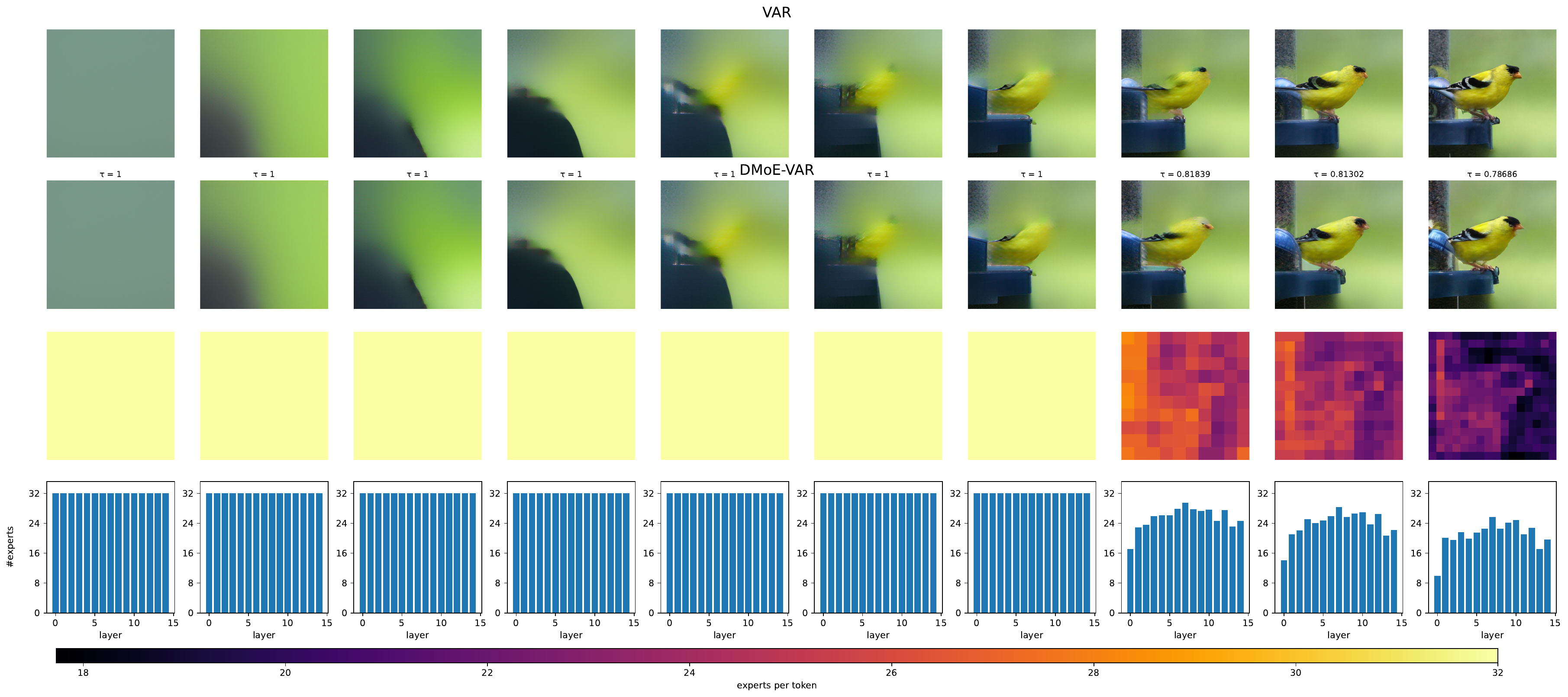}%
    \caption{Distribution of experts.}
    \label{fig:dist_experts}
  \end{subfigure}
  \vspace{-0.5em}
  \caption{\textbf{Quality–efficiency trade-off and expert routing behaviour.}
    \textbf{(a)} Qualitative comparison of DMoE-VAR and VAR samples. \textbf{(b)} Expert routing and generation patterns, shown across scales. \emph{First and Second row:} generated images from the VAR baseline and DMoE-VAR. \emph{Third row:} heat-maps of total experts allocated per token across layers. \emph{Fourth row:} bar plots of the average number of experts used at each scale.}
    \label{fig:mainresults}
\end{figure}

\textbf{Main Results.} Our method reduces FLOPs by 19\% and inference time by 11\%, while keeping FID within 1\% of the dense VAR baseline. These gains result from applying MoE layers only at the last three scales, allowing lower $\tau$ thresholds without losing image quality. We first qualitatively compare our new architecture to the VAR baseline in Figure \ref{fig:imgcomparaisons}. Images from both methods appear identical, highlighting our method’s strength in preserving semantics with identical seeds. Second, in Figure \ref{fig:dist_experts} we examine generation throughout the scales. The VAR baseline is matched by keeping scales 1–7 dense. At scale 8, the router activates many experts per token to refine uncertain regions. By the last scale, it ignores settled background and triggers only a few experts around fine edges and textures. This progressive narrowing substantially reduces both FLOPs and time while consistently maintaining high overall image quality (additional comparisons and router maps are in Appendix~\ref{section:visualAnalysis}).

\begin{figure}[htbp]
  \centering
  \begin{minipage}[t]{0.32\textwidth}
    \centering
    \includegraphics[width=\linewidth]{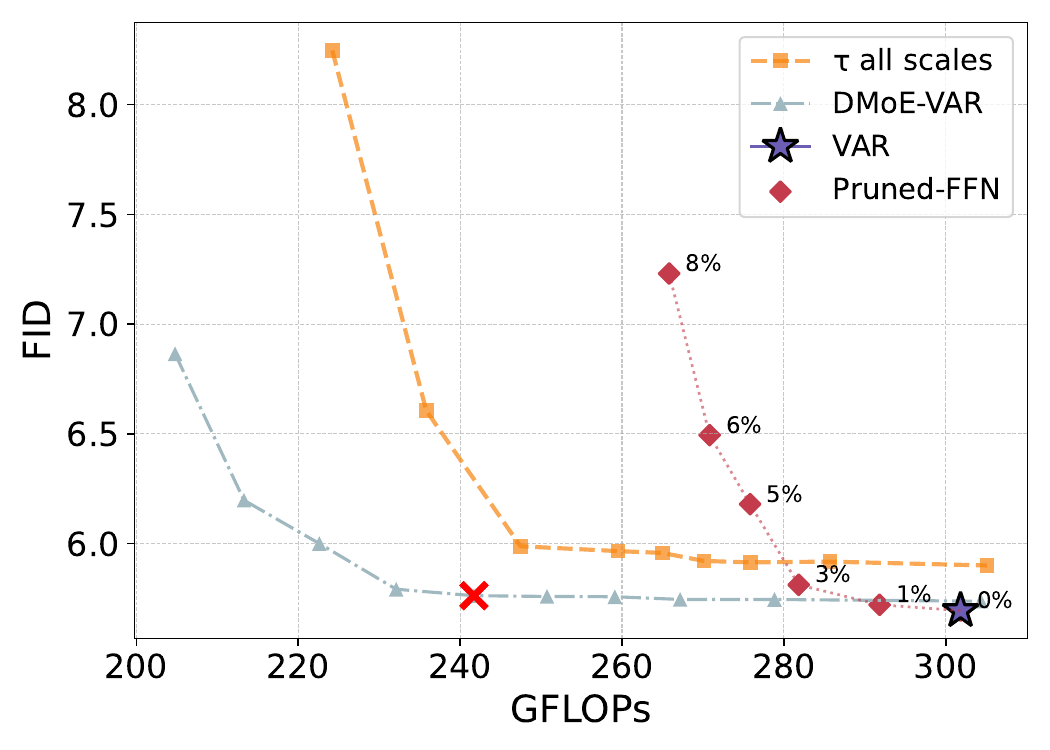}
    \captionof{figure}{FID (↓) vs GFLOP (↓). Different optimization methods utilized with VAR.}
    \label{fig:pruning}
  \end{minipage}\hfill
  \begin{minipage}[t]{0.32\textwidth}
    \centering
    \includegraphics[width=\linewidth]{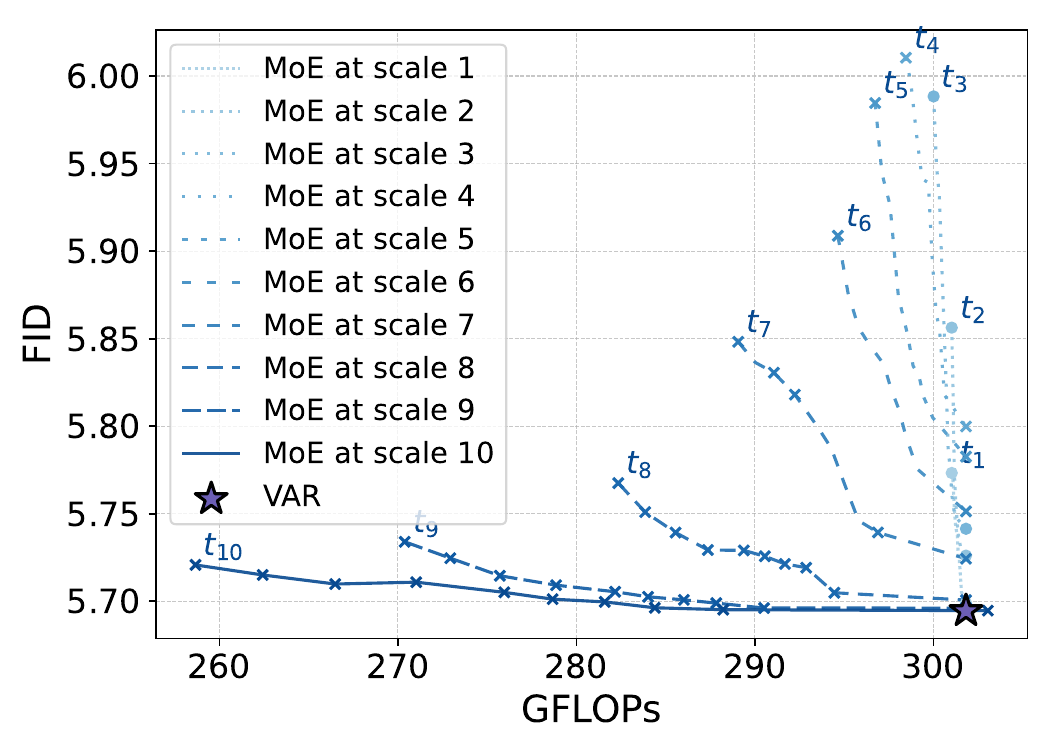}
    \captionof{figure}{FID (↓) vs GFLOP (↓). MoEfication applied to a single scale.}
    \label{fig:switch}
  \end{minipage}\hfill
  \begin{minipage}[t]{0.32\textwidth}
    \centering
    \includegraphics[width=\linewidth]{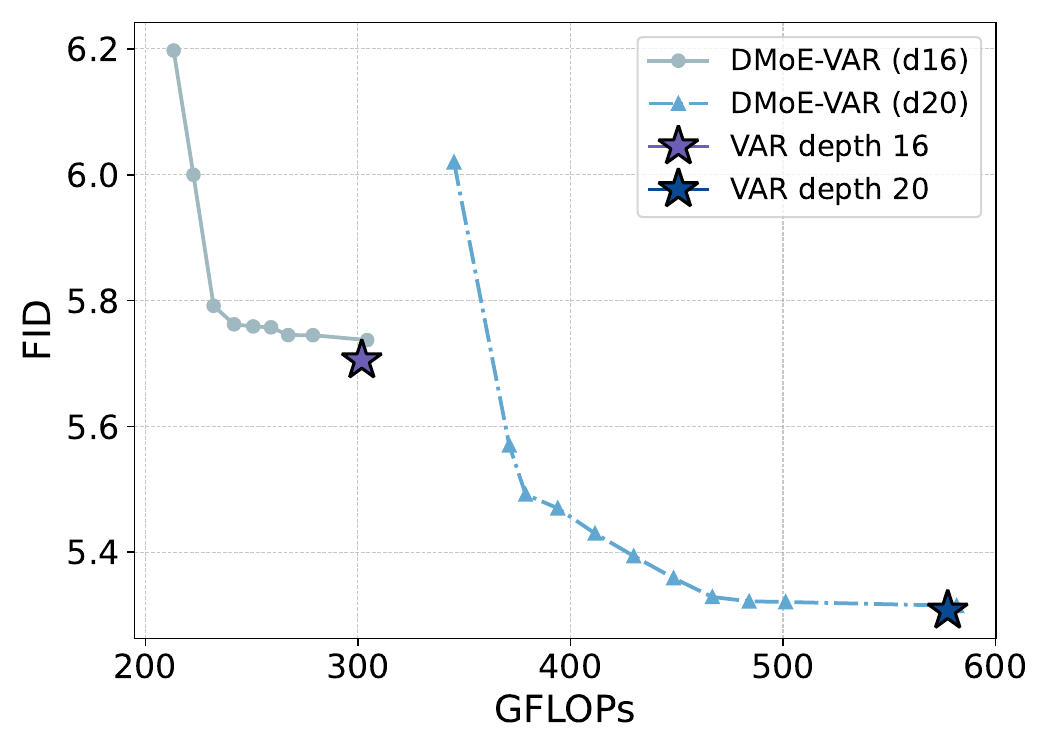}
    \captionof{figure}{FID (↓) vs GFLOP (↓). DMoE-VAR integrated into different model depths.}
    \label{fig:scling}
  \end{minipage}
\end{figure}

\textbf{$\Large \tau$-selection} In Figure~\ref{fig:pruning}, ``DMoE-VAR" employs a fixed $\tau$ for the last three scales. Varying $\tau$ produces a FID to FLOP trade-off curve, with the red cross denoting the configuration used in Figure~\ref{fig:mainresults}. This thresholding is applied post-training, enabling flexible compute allocation at inference. Applying a uniform threshold to all scales (``$\tau$ all scales”) performs worse in image quality and efficiency as errors from early scales propagate and low token counts fail to amortize the expert-loading overhead. As a result, the main gains come from sparsifying the later scales. To validate that token-wise routing is still required, we introduce ``Pruned-FFN", which prunes low-activation weights in the last three scales. The sharp performance drop confirms the necessity of our dynamic thresholding.

\textbf{Ablation Study.}  In Figure \ref{fig:switch}, we analyse the sensitivity of different scales to sparsification by selectively applying our methodology at individual scales and varying the threshold $\tau$. Scales 8–10 display the best FID to FLOP trade-off validating our targeted sparsification strategy at the last three stages. Finally, Figure~\ref{fig:scling} illustrates that our method benefits from increased model depth. Comparing the fixed-$\tau$ strategy across architectures, the FLOP/FID curve of the depth 20 model outperforms the depth 16 one. We attribute this to deeper models containing more redundant computation, which our dynamic approach can exploit to achieve greater FLOP savings.

\section{Conclusions}

We introduce a dynamic Mixture-of-Experts mechanism into Visual Autoregressive Models (VAR) to address the computational inefficiency of dense scale-wise decoding. Our scale-aware sparsification adaptively selects experts per token and resolution, exploiting redundancy in both domains to cut FLOPs while preserving image quality. Fine-scale representations show high spatial redundancy, enabling more aggressive sparsification without loss. The proposed adaptive thresholding strategy enables this selective computation, adjusting dynamically to the content and complexity of each image during generation. Future work includes fine-tuning only the last three scales for targeted sparsity and developing routing that adapts better to semantic classes.

\bibliographystyle{iclr2026_delta}

\newpage
\appendix

\section{Efficient Inference in VAR}\label{section:efficientInference}

To improve VAR’s efficiency, ScaleKV and Collaborative Decoding (CoDe) split the scale architecture into a ``drafter" and a ``refiner" stage. The drafter is responsible for generating a coarse, low-resolution version of the output sequence, while the refiner incrementally improves this draft to achieve high-quality results. ScaleKV reduces memory overhead by mapping transformer layers to these roles and compressing the key-value caches at higher resolutions \cite{li2025scalekv}. CoDe similarly adopts a two-stage decoding structure, using a large-capacity drafter for coarse representations and a lightweight refiner for fine adjustments, enabling more targeted and efficient computation \cite{chen2024code}. These architectures show that dividing decoding into specialized components can yield substantial efficiency gains without compromising output quality \cite{xie2024litevar, gong2025}.

\section{Hyper-parameter Table}\label{Appendix:Hyperparameter}

\begin{table}[ht]
\centering
\resizebox{0.94\textwidth}{!}{%
\begin{tabular}{lll}
\toprule
\textbf{Hyper-parameter} & \textbf{Value} & \textbf{Description}\\
\midrule
\multicolumn{3}{c}{\textbf{Sparsification}} \\
\midrule
    Batch Size & 512 & Global batch size\\
    Epochs & 2 & Training epochs\\
    Loss Type & \texttt{ce} & Cross-entropy loss\\
    Optimizer & \texttt{adam} & Adam Optimizer\\
    Learning Rate & $2\times10^{-5}$ & Learning rate\\
    Weight Decay & 0.05 & Regularization\\
    Scheduler & \texttt{linear} & Learning Rate Scheduler\\
    Warmup Steps & 0.2 epochs & Initial learning rate warm-up\\
    Gradient Norm Clip & 1.0 & Gradient clipping norm\\
    \midrule
    \multicolumn{3}{c}{\textbf{Expert Split}} \\
    \midrule
    Expert Size & 128 & Hidden dimension of each expert\\
    Experts Class & \texttt{execute\_all} & Expert selection mode\\
    Activation & GeLU & Activation function between Experts\\
\midrule
\multicolumn{3}{c}{\textbf{Router Training}} \\
\midrule
    Epochs & 2 & Training epochs\\
    Batch Size & 256 & Batch size\\
    Learning Rate & $1 \times 10^{-3}$ & Optimizer learning rate\\
    Router Loss Type & \texttt{mse} & Mean Squared Error\\
    Router Depth & 2 & Hidden layers in router\\
    Router Width & 128 & Width of router hidden layers\\
    Activation & \texttt{gelu} & Router internal activation\\
    Output Activation & \texttt{abs} & Router output activation\\
    Labels Norm & 2 & Label normalization\\
    Number of Experts & 128 & Experts width per MoE layer\\
\midrule
\multicolumn{3}{c}{\textbf{Evaluation (FID/IS)}} \\
\midrule
    Batch Size & 128 & Batch size during sampling\\
    Tau & [0.81839, 0.81302, 0.78686] & Tau per scale\\
    Tau as List & True & Interpret tau as list\\
    Expert Index Switch & 7 & Scale to start replacing with MoE\\
    CFG & 1.5 & Classifier-free guidance\\
    Top-P & 0.96 & Nucleus sampling probability\\
    Top-K & 900 & Sampling truncation parameter\\
    Forward Mode & \texttt{dynk\_max} & Expert selection mode\\
    Samples per Class & 10 & Number of images per class\\
    Random Seed & 0 & Deterministic sampling\\
    TF32 Enabled & True & TensorFloat32 usage\\
\bottomrule
\end{tabular}%
}
\caption{Hyperparameter Settings for all Experiments}
\label{tab:hyperparams_dsti}
\end{table}
\newpage

\section{Design Exploration}\label{Section:DesignExploration}

To mitigate errors from the router, we employ an ``oracle" router strategy in Sections \ref{Section:Sparsity} and \ref{Section:MoEfication}. Specifically, tokens are passed through the expert modules twice: first, to compute the activation norms from \eqref{routernorm} and, then, to apply dynamic thresholding based on determined $\tau$ values (Eq. \eqref{dynamicthresh}). We exclude the first pass from the fvcore measurements to obtain realistic FLOP estimates.

\subsection{Sparsity}\label{Section:Sparsity}

\begin{figure}[htbp]
  \centering
  %––– Cross‐entropy loss (left) –––––––––––––––––––––––––––––––––
  \begin{subfigure}[t]{0.32\textwidth}
    \includegraphics[width=\linewidth]{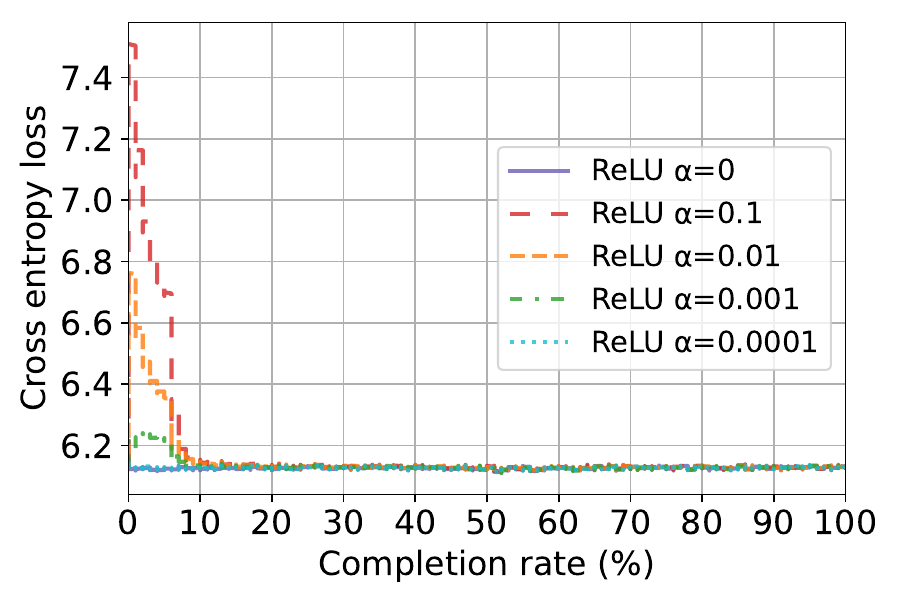}
    \caption{Cross‐entropy loss (↓)}
    \label{fig:ce_loss}
  \end{subfigure}\hfill
  %––– Hoyer‐loss (center) –––––––––––––––––––––––––––––––––––
  \begin{subfigure}[t]{0.32\textwidth}
    \includegraphics[width=\linewidth]{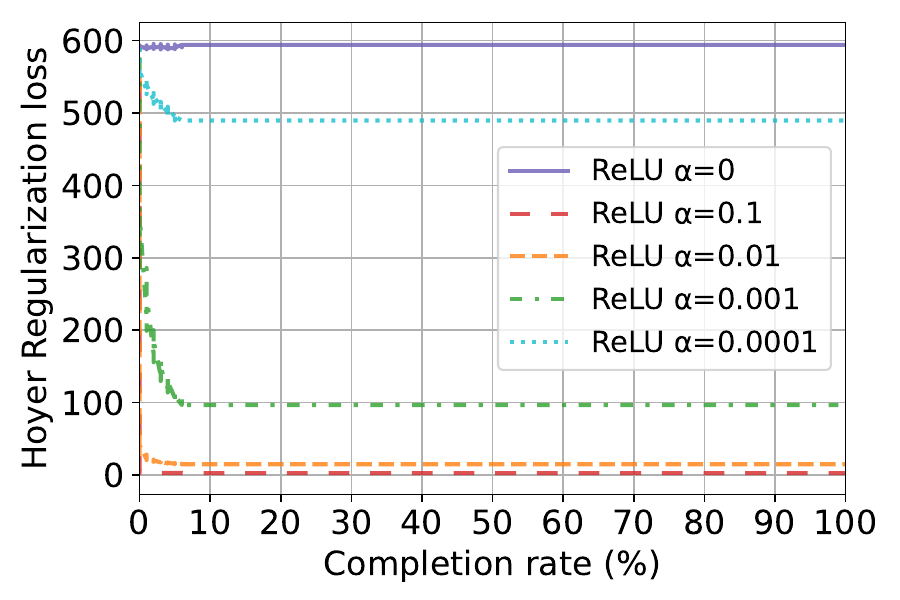}
    \caption{Hoyer regularization loss}
    \label{fig:hoyer_loss}
  \end{subfigure}\hfill
  %––– FID vs Experts (right) ––––––––––––––––––––––––––––––––
  \begin{subfigure}[t]{0.32\textwidth}
    \includegraphics[width=\linewidth]{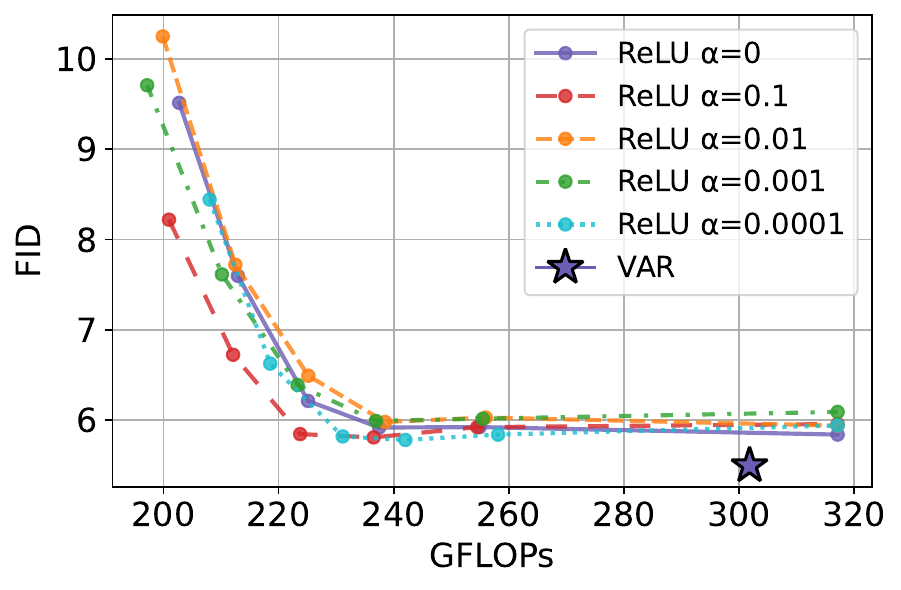}
    \caption{FID (↓) vs GFLOPs (↓)}
  \end{subfigure}
    \caption{Training curves for our model fine-tuned under the combined loss 
      $L = L_{\mathrm{CE}} + \alpha\,L_{s}$ (see Eq.~\ref{eq:losses}), where $L_{s}$ is the Hoyer sparsity penalty, plotted for five different values of $\alpha$ over two epochs. For FID evaluation, we created 512 experts of size 8 using the oracle routing strategy.}
  \label{fig:all_three}
\end{figure}

\begin{wrapfigure}{r}{0.33\textwidth}
  \centering
  \includegraphics[width=\linewidth]{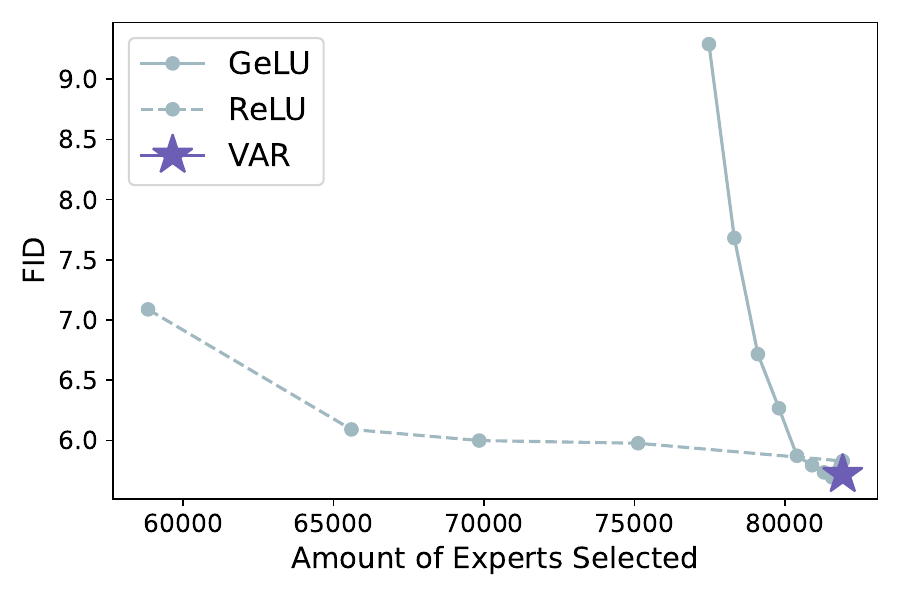}
  \caption{Comparison between the GeLU, ReLU and VAR models}
  \label{fig:gelu}
\end{wrapfigure}

We fine-tune models using the joint loss $L(x) = L_{\mathrm{CE}}(\hat{y}, y) + \alpha L_{s}(x)$ (eq.~\eqref{eq:losses}). As shown in Figures \ref{fig:ce_loss} and \ref{fig:hoyer_loss}, convergence is reached within roughly one-fifth of an epoch, requiring minimal additional fine-tuning steps. While the resulting models achieve comparable cross-entropy and FID scores, they differ in sparsity loss. We believe this happens because the sparsity regulariser mostly affects the early, complex scales, which have little impact on total FLOPs, while the later scales make up for the errors from earlier layers as they contain far more tokens. As a result, variations in sparsity translate into minimal FLOP savings. This points to ReLUfication as the dominant source of sparsity, a claim supported by our ablation without ReLUfication (Figure \ref{fig:gelu}), where performance degrades quickly, showing Hoyer regularisation alone is insufficient for robust sparsity.

\subsection{MoEfication and Efficiency}\label{Section:MoEfication}

\begin{figure}[h]
  \centering
  \begin{subfigure}[b]{0.48\textwidth}
    \includegraphics[width=\linewidth]{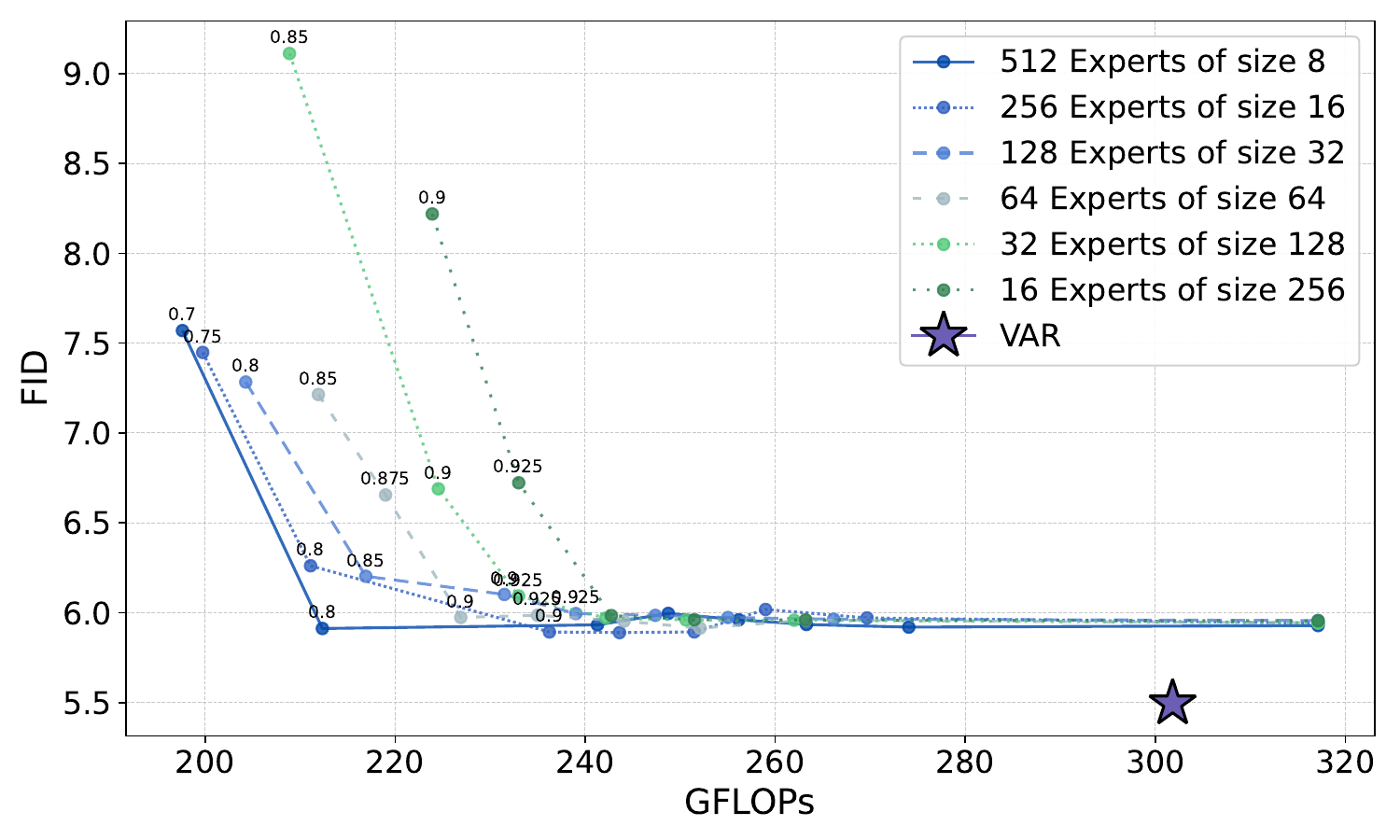}
    \caption{Different Experts Sizes}
    \label{MoeficationPlot}
  \end{subfigure}
  \hfill
  \begin{subfigure}[b]{0.48\textwidth}
    \includegraphics[width=\linewidth]{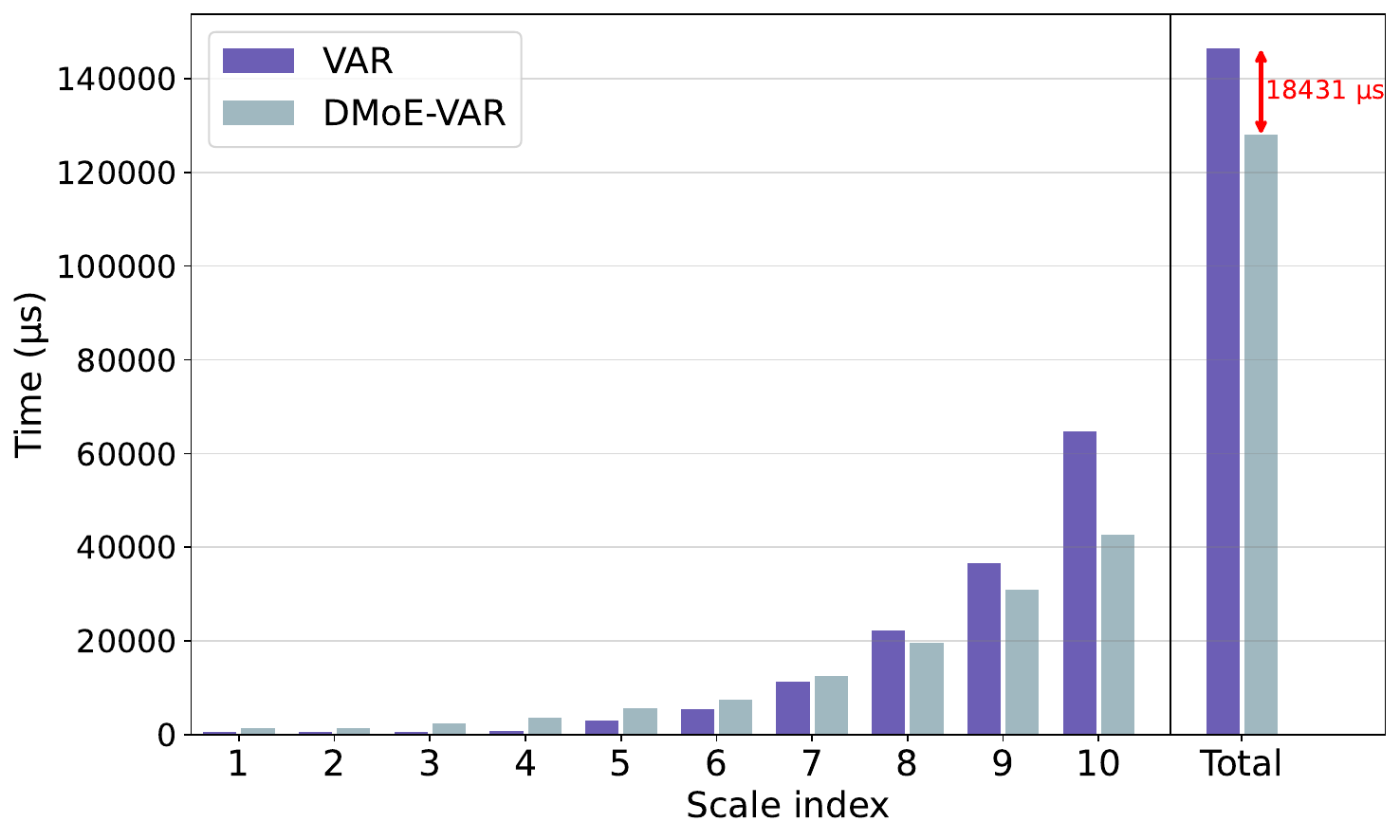}
    \caption{Efficiency of VAR vs DMoE-VAR}
    \label{Timeplotsfore128}
  \end{subfigure}
  \vspace{-1ex}
  \caption{\textbf{Expert sizes vs. compute and speed.} For the sparsified model fine-tuned under the combined loss \(L = L_{\mathrm{CE}} + \alpha\,L_{s}\) (Eqs.~\ref{eq:losses}, \(\alpha=0.1\)) using the oracle routing strategy.
    \textbf{(a)} FID (↓) vs.\ GFLOPs (↓) for various expert sizes and \(\tau\) values (annotated).  
    \textbf{(b)} Wall-clock time per scale comparing VAR and DMoE-VAR (32×128 experts, \(\tau=0.7\)), showing combined CPU + GPU time for batch size 128.}
\end{figure}
\newpage

To leverage sparse MoE architectures effectively, we evaluate the trade-off between expert granularity and performance (Figure \ref{MoeficationPlot}). The model achieves finer-grained representations as the number of experts increases, which is reflected in lower FID scores. However, due to GPU threading overhead for each expert, expert sizes of at least 128 are required to obtain runtime benefits. As a result, while a configuration of 512 experts of size 8 yields the best FID, practical deployment forces us to select the 32 experts of size 128, striking a balance between FID and speed-up.

\begin{figure}[!ht]
\centering
\begin{subfigure}[b]{0.49\linewidth}
\centering
\includegraphics[width=\linewidth]{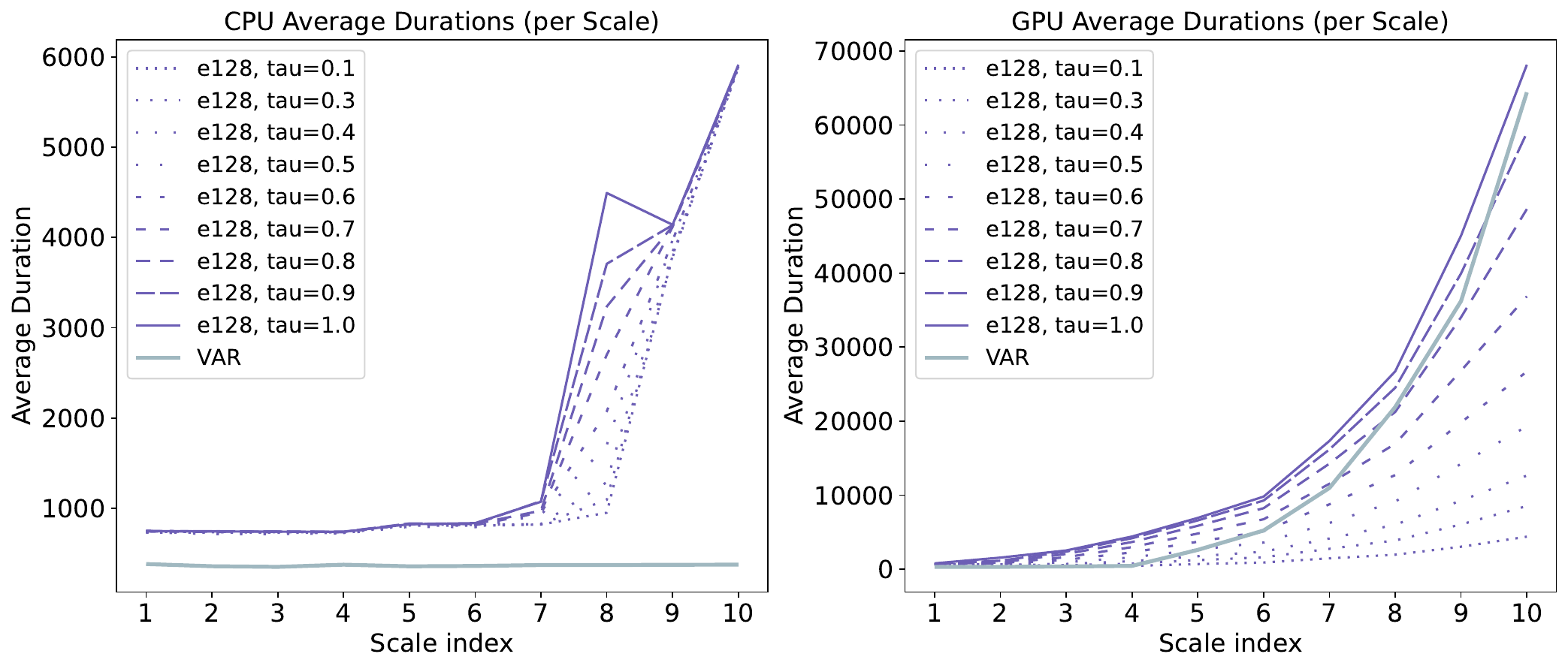}
\caption{CPU and GPU time for 128 experts width.}
\label{fig:moe_tau_128}
\end{subfigure}
\hfill
\begin{subfigure}[b]{0.49\linewidth}
\centering
\includegraphics[width=\linewidth]{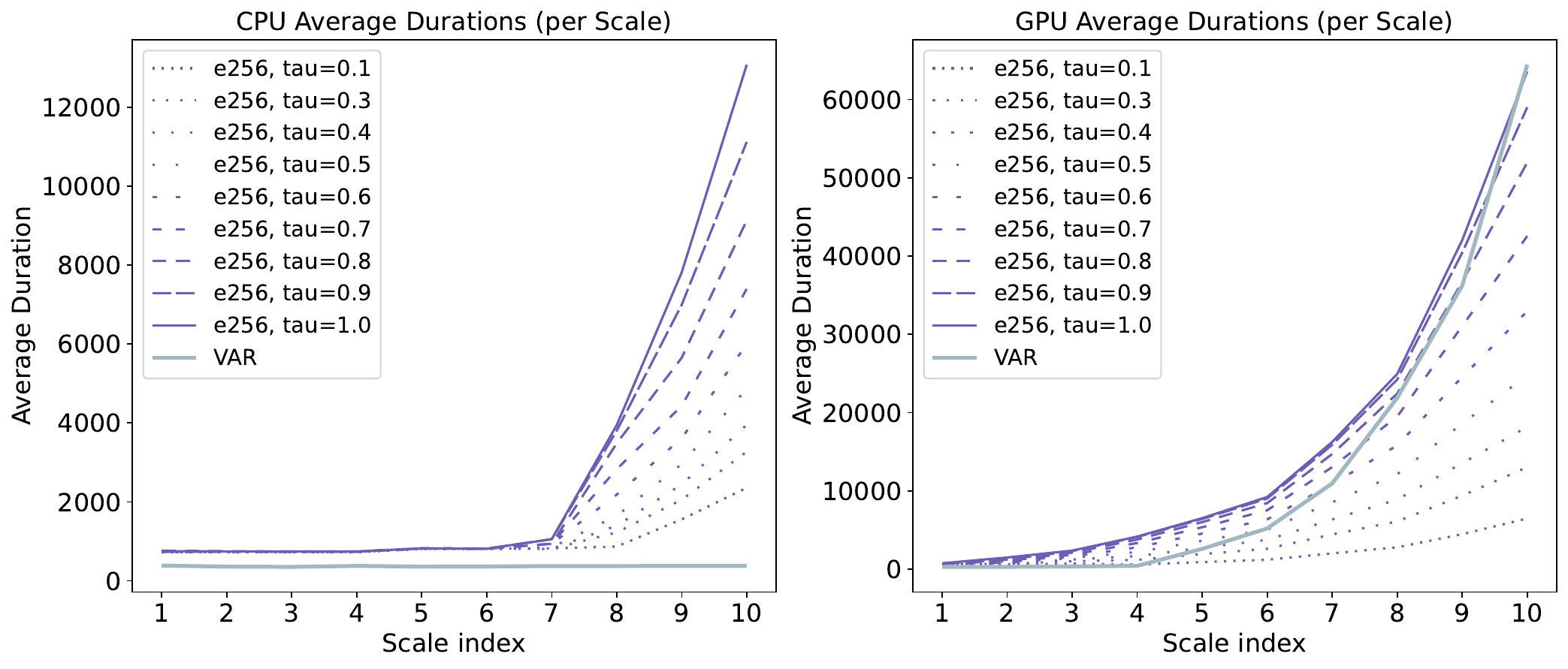}
\caption{CPU and GPU time for 256 experts width.}
\label{fig:moe_tau_256}
\end{subfigure}
\caption{Running time for a single batch under the $\tau$ threshold rule.}
\label{fig:moe_tau_comparison}
\end{figure}

It is important to note that expert size is not the only factor driving efficiency optimisations. Other parameters, such as the number of active experts (regulated by $\tau$), CPU/GPU utilization, batch size and hardware specifications, all influence runtime. Figure \ref{fig:moe_tau_comparison} reports the wall clock time on the CPU and GPU for a single batch using the $\tau$ selection in \eqref{dynamicthresh}. For a fixed batch size the GPU becomes faster when expert width grows, because each activation touches fewer experts. The CPU trend is reversed due to sorting of the experts for each token. Throughout the experiments we push the batch size until the model almost reaches an out of memory (OOM) event. On our hardware, a batch size of $128$ is the largest configuration that fits. With this batch size, any smaller experts would result in an OOM, and increasing the batch size to $256$ fails to run for any expert size. It is to note that this set‑up is optimal for our machine, other systems will require different values. 

In our most realistic setup, we compare the throughput of the VAR and DMoE-VAR model using 128-sized experts with a threshold of $\tau = 0.7$ (Figure \ref{Timeplotsfore128}). Since GPU speed-ups scales with the amount of tokens, only the later scales (8–10) display meaningful acceleration, resulting in an overall runtime improvement of approximately 12\%. These results underscore two practical guidelines: (i) fewer, larger experts are preferable for end-to-end efficiency, and (ii) MoEfication should only be applied on deeper scales where enough tokens are present to amortize the threading overhead.

\subsection{Router selection}\label{Section:RouterAndTau}

\begin{figure}[H] 
  \centering
  % first panel
  \begin{subfigure}[b]{0.49\textwidth}
    \includegraphics[width=\linewidth]{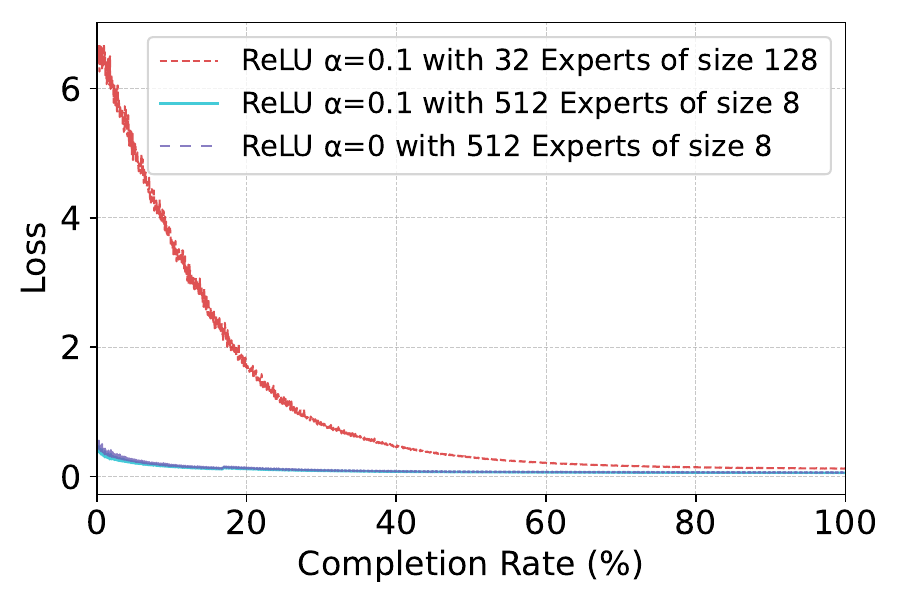}
    \caption{Router Loss}
    \label{fig:rl}
  \end{subfigure}
  \hfill
  % second panel
  \begin{subfigure}[b]{0.49\textwidth}
    \includegraphics[width=\linewidth]{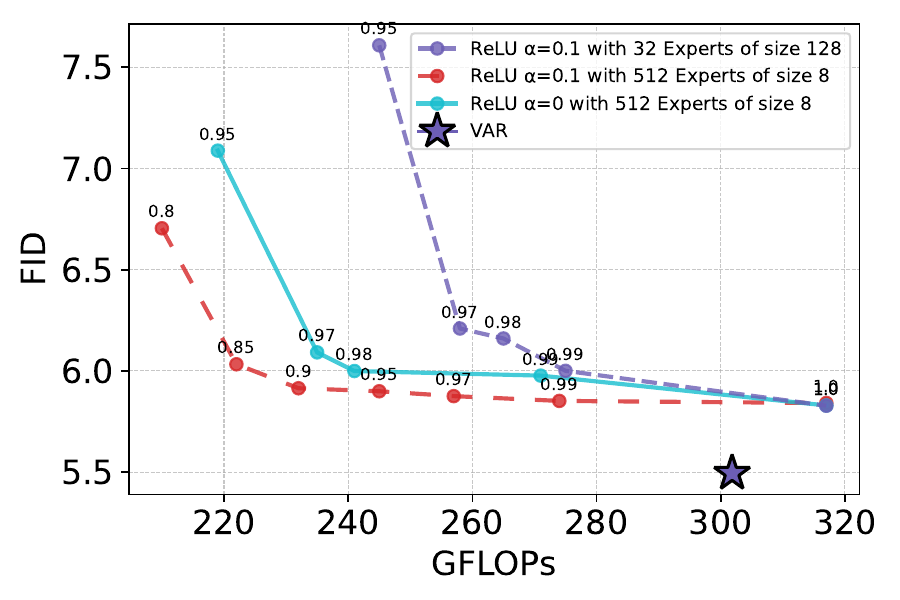}
    \caption{FID for different routers}
    \label{fig:fr}
  \end{subfigure}
  \vspace{1ex}
  \caption{\textbf{Router analysis and hyperparameter tuning.} We optimize the router loss  
    $L_{r}(z) = \frac{1}{n} \sum_{i=1}^{n} \bigl(R(z)_i - \|E_{i}(z)\|\bigr)^2$ using the previously sparsified model  
    $L = L_{\mathrm{CE}} + \alpha\,L_{s}$ (Eqs.~\ref{eq:losses}) with \(\alpha\in\{0,0.1\}\) and two expert configurations (32 experts of size 128 and 512 experts of size 8).  
    \textbf{(a)} Router loss (↓) vs.\ completion rate over two epochs.  
    \textbf{(b)} FID (↓) vs GFLOPs (↓) for the different configurations.}
  \label{fig:three_sidebyside}
\end{figure}

Incorporating a learned router introduces additional parameters and potential prediction errors. Figures \ref{fig:rl} and \ref{fig:fr} illustrate the learned router’s loss and FID. Notably, for equivalent expert sizes, the router achieves improved FID scores when trained with sparsity regularization $\alpha=0.1$, compared to $\alpha=0$. We hypothesize that this occurs because higher sparsity regularization results in more zero-valued expert norms, simplifying the router’s prediction task and thus enhancing overall FID.

\section{Visual Analysis}\label{section:visualAnalysis}

\setlength{\tabcolsep}{0pt}
\renewcommand{\arraystretch}{0}
\newcommand{\cellwidth}{\dimexpr\linewidth/3\relax}
\begin{figure}[!ht]
  \centering
  % ------------ LEFT GRID ------------
  \begin{minipage}{0.48\linewidth}
    \centering
    \begin{tabular}{@{}ccc@{}}
      \includegraphics[width=\cellwidth]{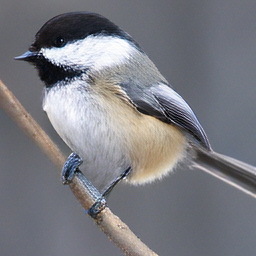} &
      \includegraphics[width=\cellwidth]{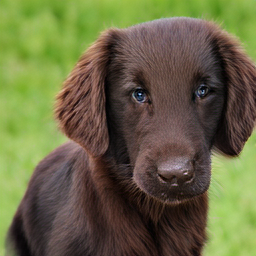} &
      \includegraphics[width=\cellwidth]{class_0076_1.png} \\%
      \includegraphics[width=\cellwidth]{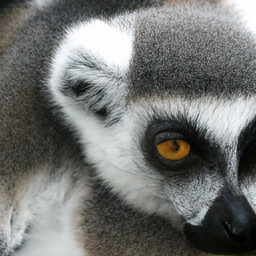} &
      \includegraphics[width=\cellwidth]{class_0385_1.png} &
      \includegraphics[width=\cellwidth]{class_0562_1.png} \\%
      \includegraphics[width=\cellwidth]{class_0985_1.png} &
      \includegraphics[width=\cellwidth]{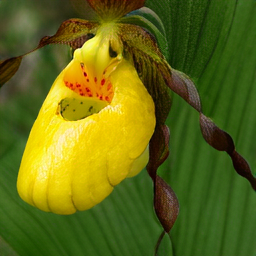} &
      \includegraphics[width=\cellwidth]{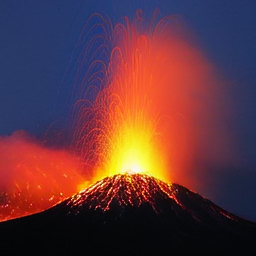} \\
    \end{tabular}
    {\small\textbf{DMoE-VAR}}
  \end{minipage}
  \hfill
  % ------------ RIGHT GRID ------------
  \begin{minipage}{0.48\linewidth}
    \centering
    \begin{tabular}{@{}ccc@{}}
      \includegraphics[width=\cellwidth]{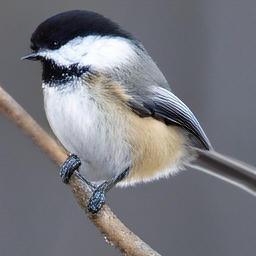} &
      \includegraphics[width=\cellwidth]{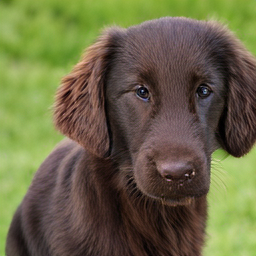} &
      \includegraphics[width=\cellwidth]{class_0076.png} \\%
      \includegraphics[width=\cellwidth]{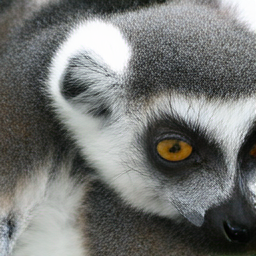} &
      \includegraphics[width=\cellwidth]{class_0385.png} &
      \includegraphics[width=\cellwidth]{class_0562.png} \\%
      \includegraphics[width=\cellwidth]{class_0985.png} &
      \includegraphics[width=\cellwidth]{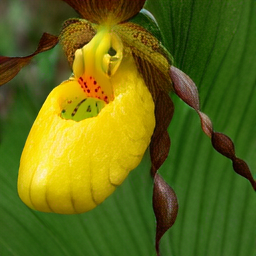} &
      \includegraphics[width=\cellwidth]{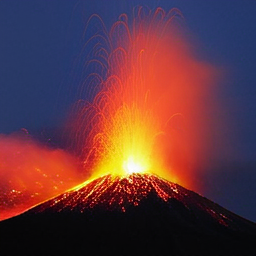} \\
    \end{tabular}
    {\small\textbf{VAR}}
  \end{minipage}

  \caption{More qualitative comparisons of DMoE-VAR and VAR samples. Generated images were produced using classifier-free guidance and Gaussian smoothing.}
  \label{fig:dual_grids_tight}
\end{figure}

\begin{figure}[H]
  \centering
  % zero \tabcolsep inside each pair, keep \quad between scale‐blocks
  \setlength{\tabcolsep}{0pt}
  \begin{tabular}{@{}cc@{} @{\quad} cc@{} @{\quad} cc@{}}
    % header
    \multicolumn{2}{c}{\textbf{Scale 8}}
    & \multicolumn{2}{c}{\textbf{Scale 9}}
    & \multicolumn{2}{c}{\textbf{Scale 10}} \\
    \midrule
    % Class 213
    \includegraphics[width=0.16\textwidth]{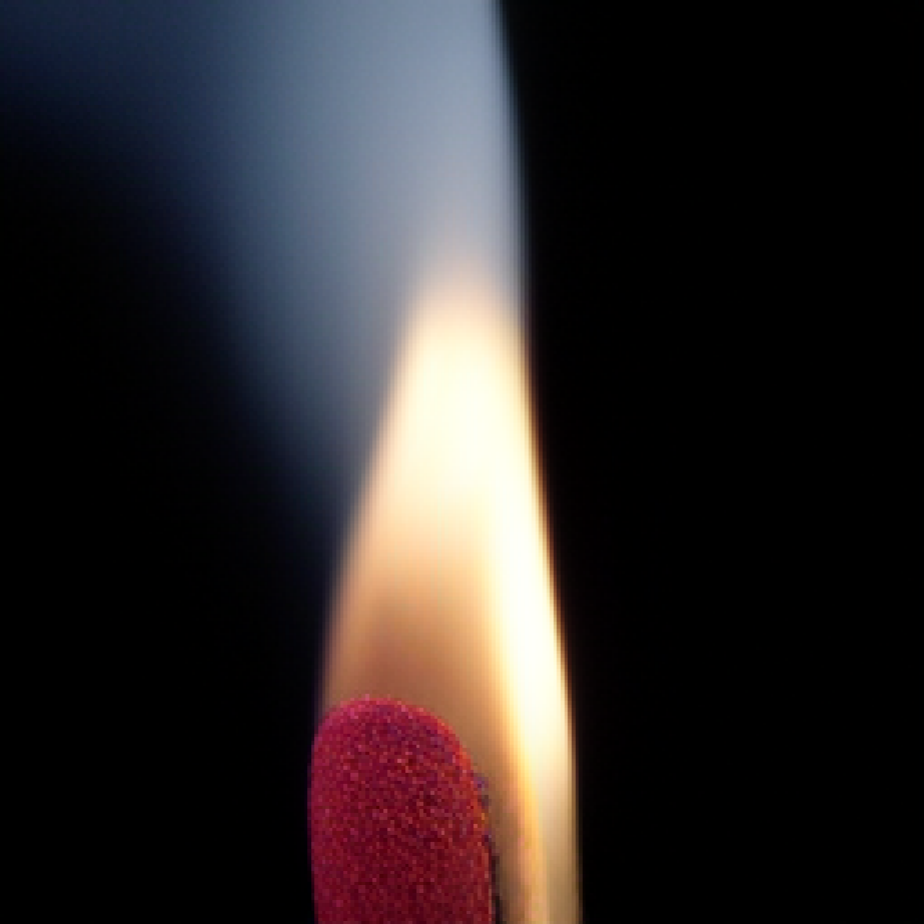}
    & \includegraphics[width=0.16\textwidth]{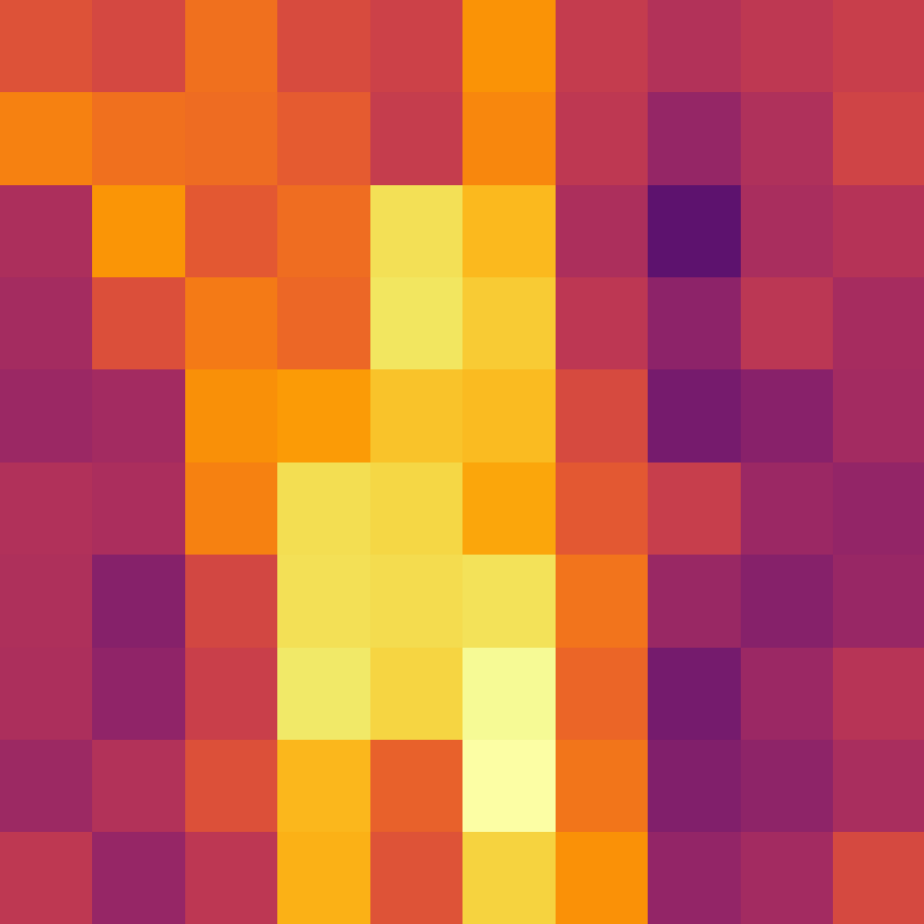}
    & \includegraphics[width=0.16\textwidth]{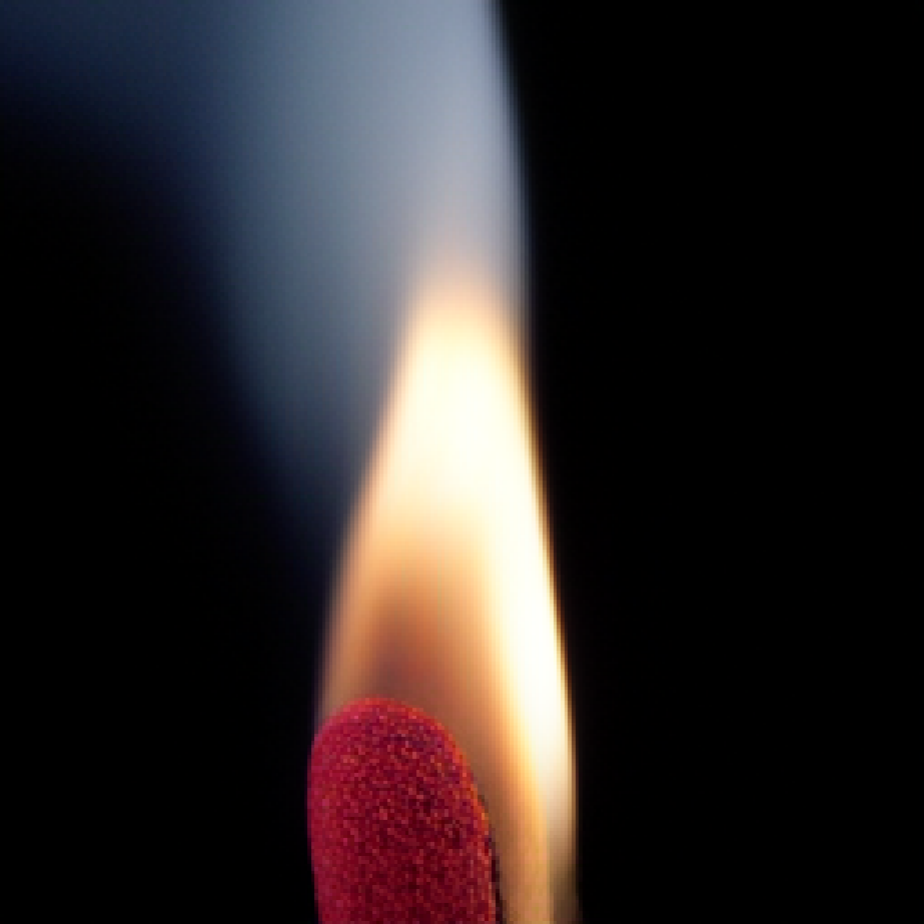}
    & \includegraphics[width=0.16\textwidth]{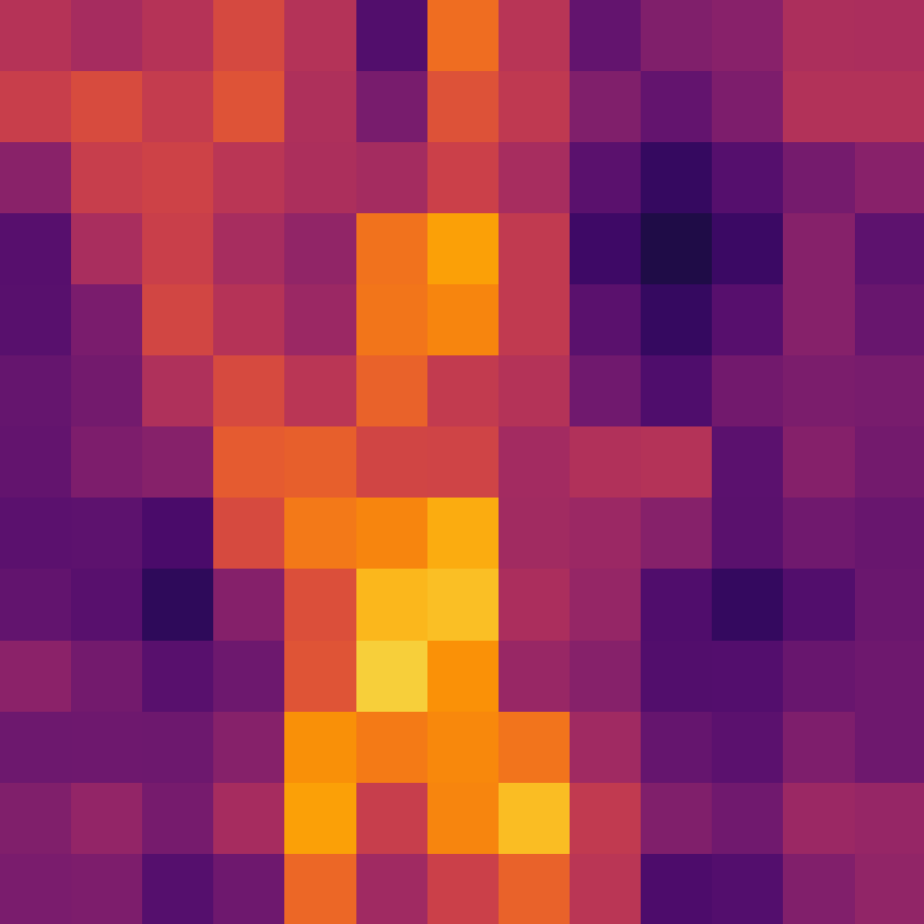}
    & \includegraphics[width=0.16\textwidth]{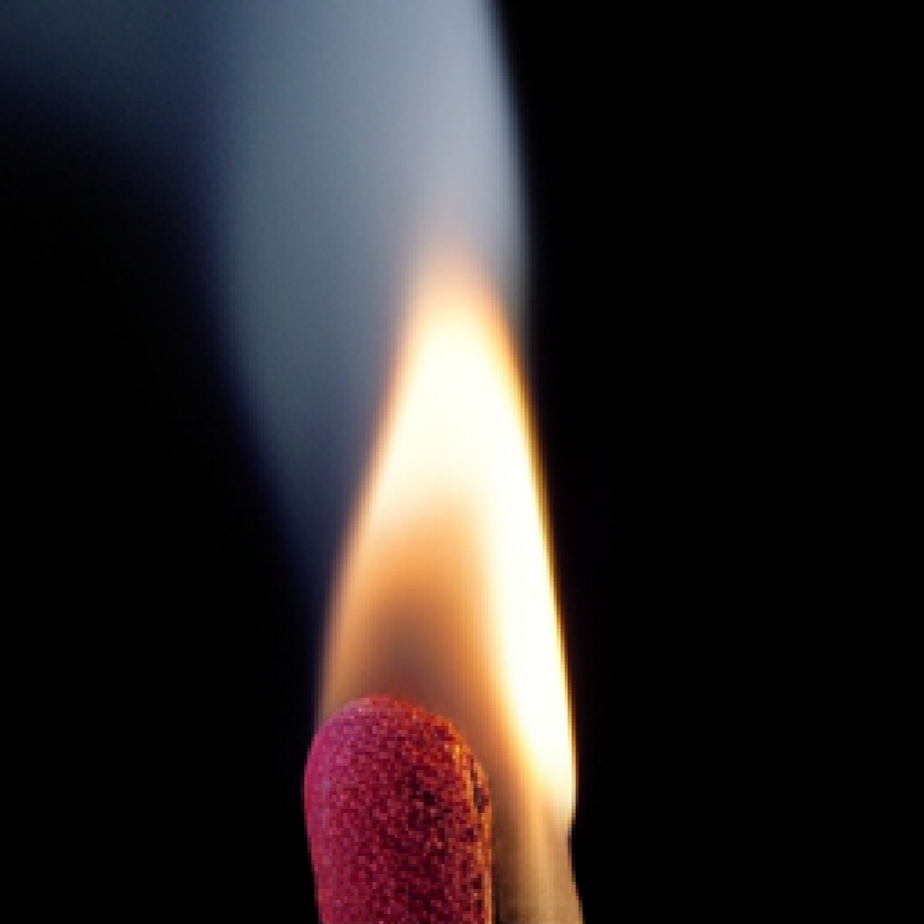}
    & \includegraphics[width=0.16\textwidth]{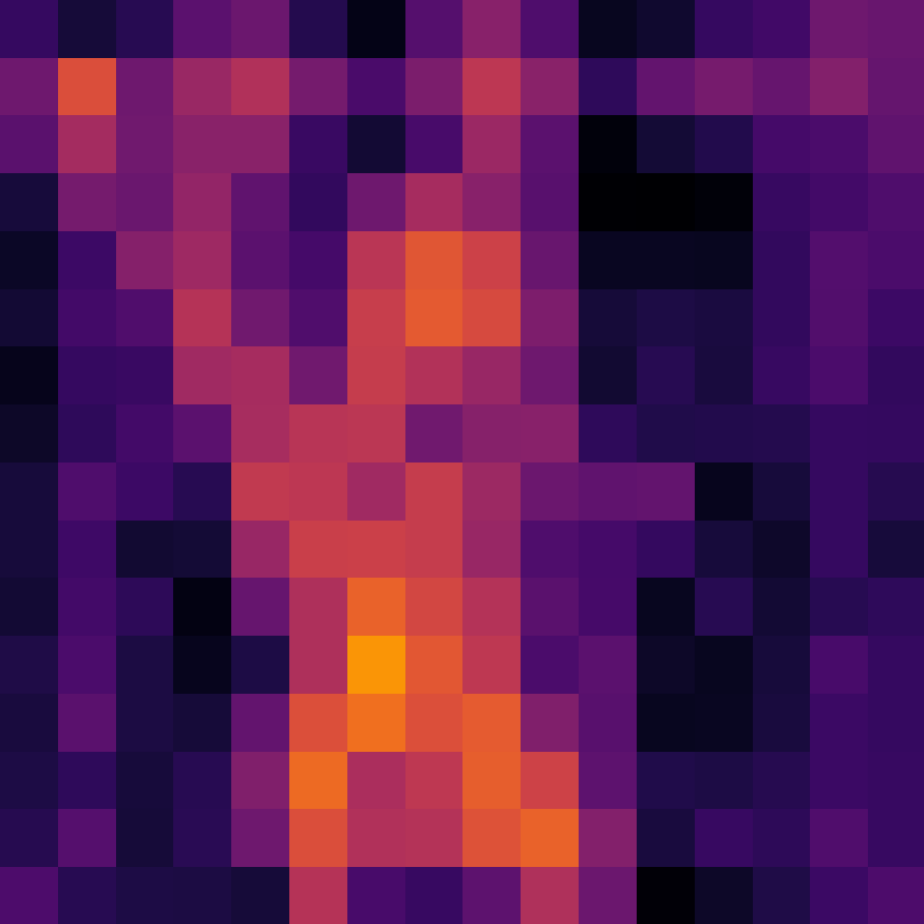}
    
    \\[1em]
    % % Class 319
    % \includegraphics[width=0.16\textwidth]{generated_scale_7_1.pdf}
    % & \includegraphics[width=0.16\textwidth]{expert_map_scale_7_1.pdf}
    % & \includegraphics[width=0.16\textwidth]{generated_scale_8_1.pdf}
    % & \includegraphics[width=0.16\textwidth]{expert_map_scale_8_1.pdf}
    % & \includegraphics[width=0.16\textwidth]{generated_scale_9_1.pdf}
    % & \includegraphics[width=0.16\textwidth]{expert_map_scale_9_1.pdf}
    % \\[1em]
    % Class 644
    \includegraphics[width=0.16\textwidth]{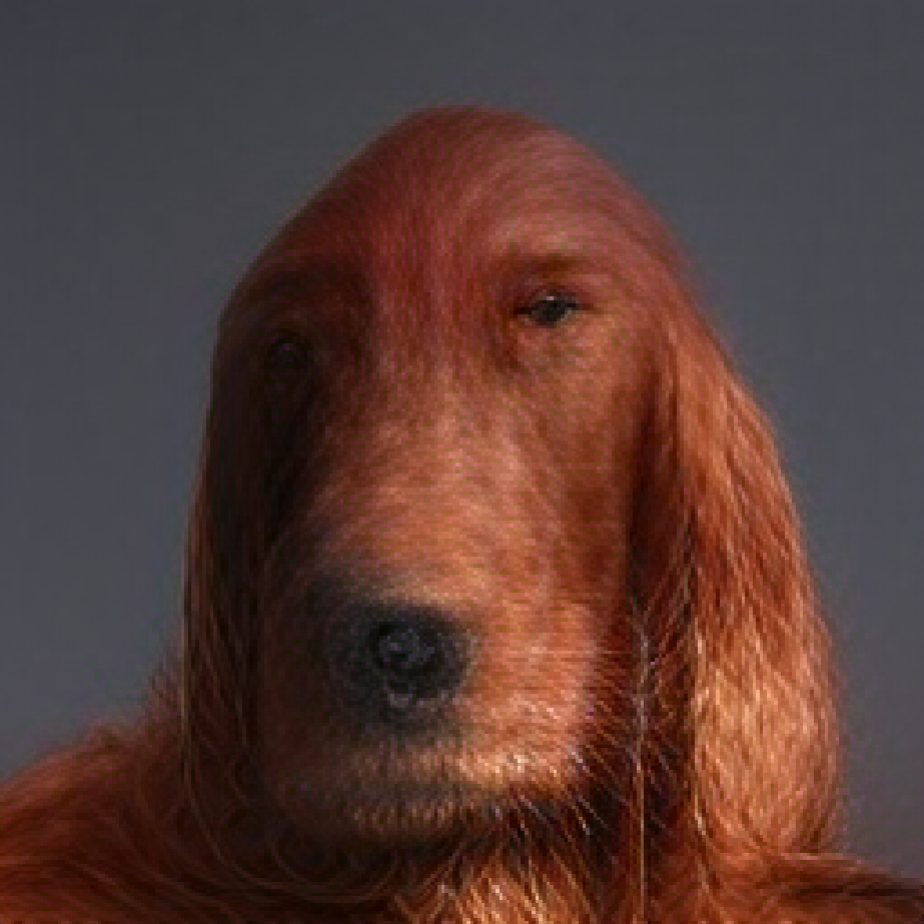}
    & \includegraphics[width=0.16\textwidth]{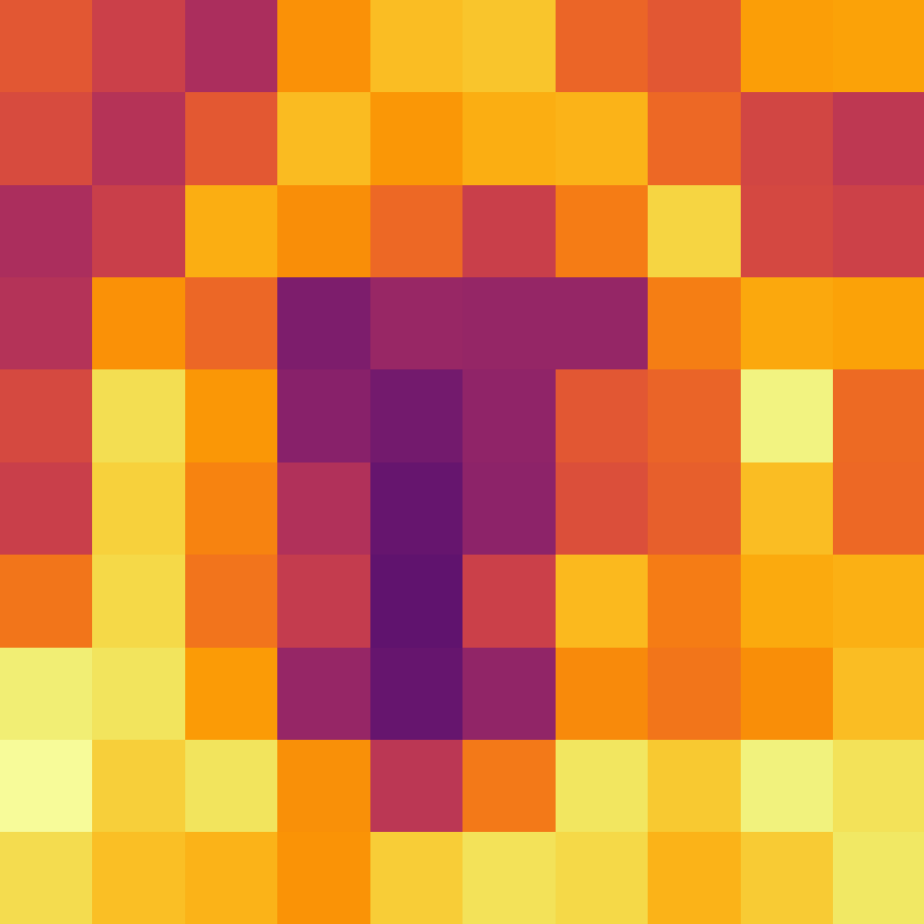}
    & \includegraphics[width=0.16\textwidth]{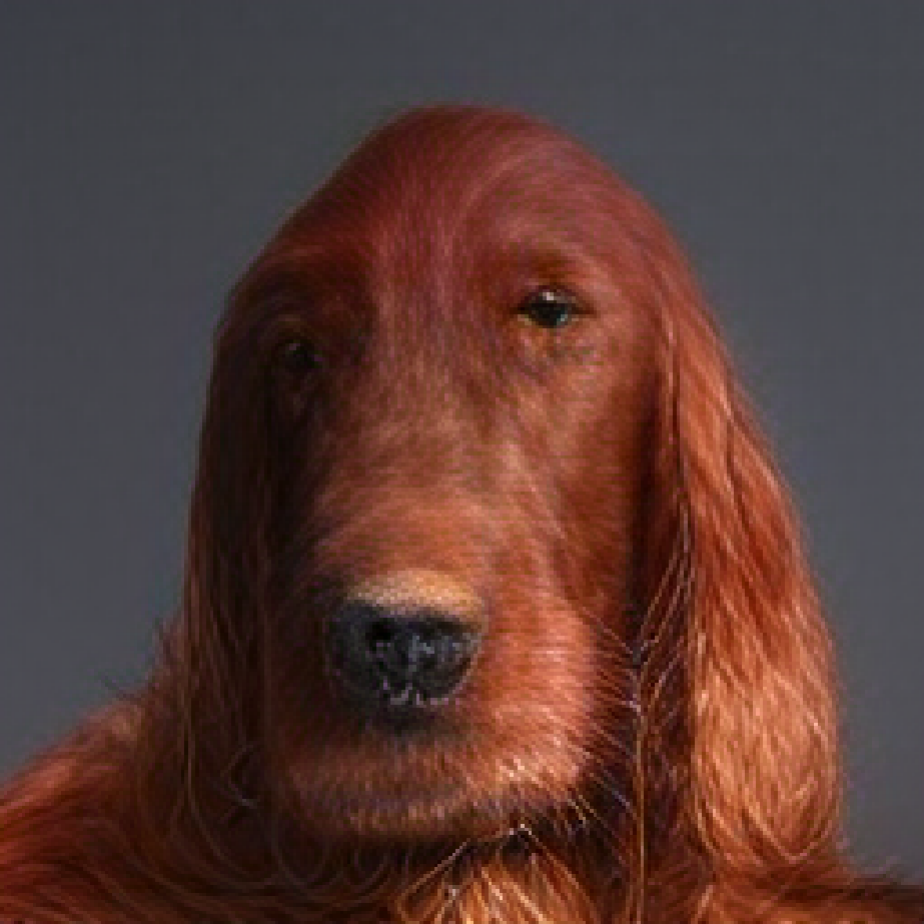}
    & \includegraphics[width=0.16\textwidth]{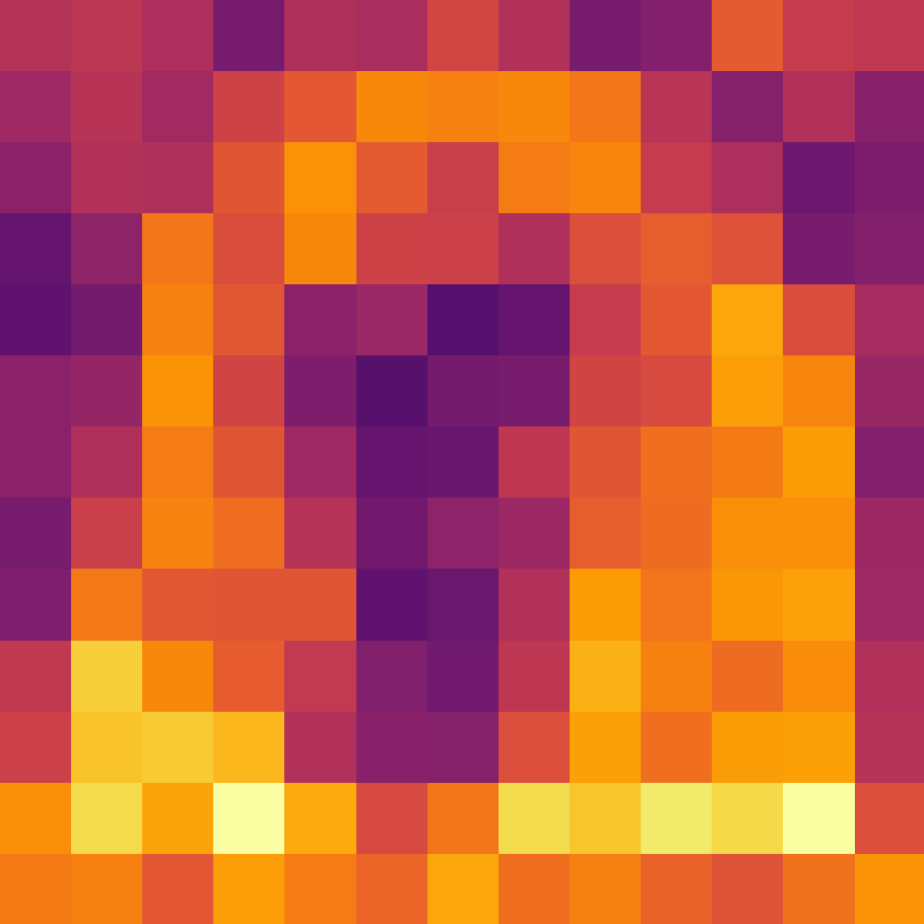}
    & \includegraphics[width=0.16\textwidth]{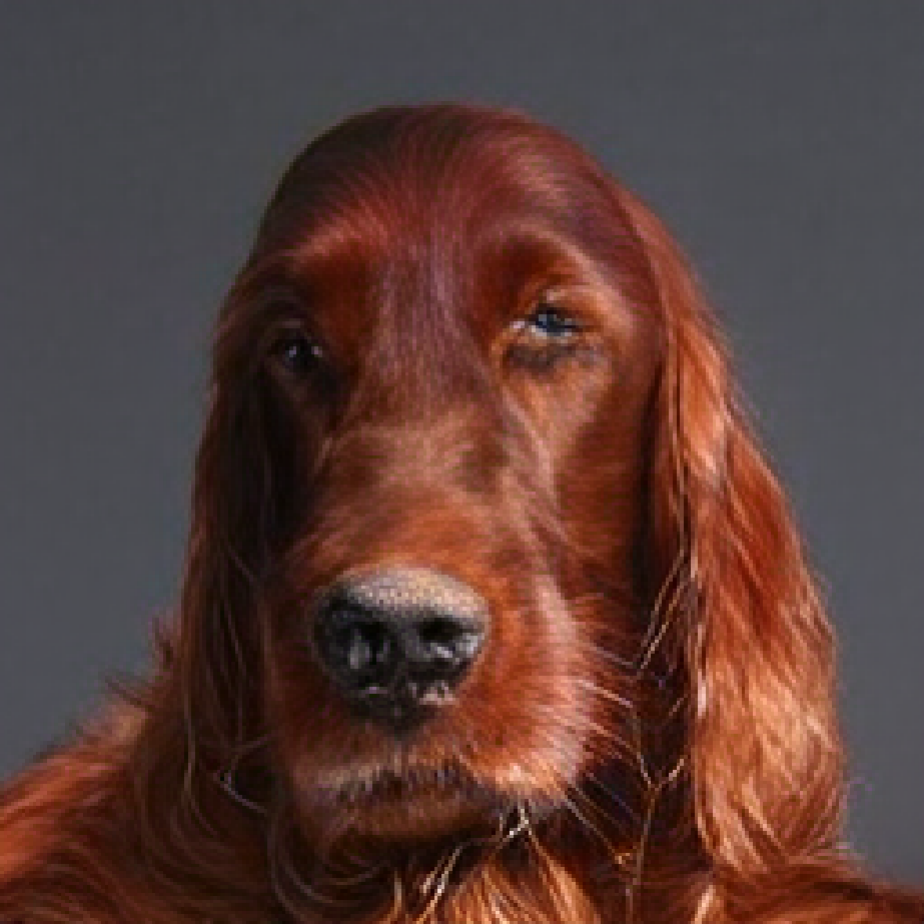}
    & \includegraphics[width=0.16\textwidth]{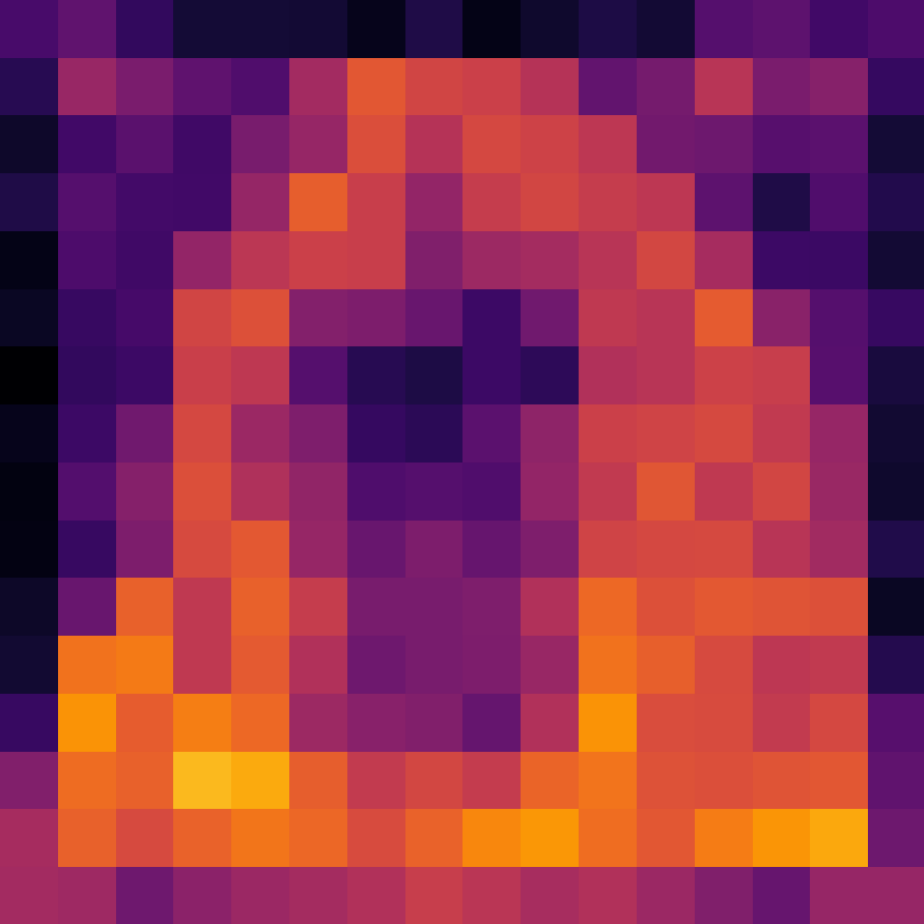}
    \\
    \bottomrule
  \end{tabular}
    \caption{At each scale (8, 9, and 10), the generated image (left) is paired with its corresponding expert allocation map (right), obtained by summing the activated experts per token across both conditioned and unconditioned samples.}
  \label{fig:sparsity_grid}
\end{figure}

Figure \ref{fig:sparsity_grid} shows how DMoE-VAR allocates experts at scales 8 to 10 as $\tau$ decreases. At scale 8, the sparsity map reflects higher uncertainty and spreads compute across the whole image. As resolution grows finer, the experts are localised to the most important regions, separating the pattern into foreground and background. However, the dog’s nose still triggers only a few experts, even if it lies at the centre of the image. We hypothesise that, with large expert width, just a small set of experts dominate the norms in such specialised regions, suppressing the others. While this focused allocation preserves global fidelity, it can miss very fine details compared to the dense VAR baseline. For example, the spider’s eyes in Figure \ref{fig:dual_grids_tight} are missing with our model. 

\end{document}